\documentclass[journal,twoside]{IEEEtran}

\usepackage[utf8]{inputenc}
\usepackage{tikz}
\usepackage{subfigure}
\usepackage[noend]{algpseudocode}
\usepackage{pgfplots}
\usepackage{algorithm}
\usepackage{upgreek}
\usepackage{stackengine}
\ifCLASSINFOpdf
\else
\fi
\usepackage{amsmath}
\usepackage{amssymb}
\usepackage{array}
\usepackage{multirow}
\pgfplotsset{compat=1.13}
\begin{document}

\title{Deep Reinforcement Learning for Haptic Shared Control in Unknown Tasks}
%
%
%

\author{Franklin~Cardeñoso~Fernandez
        and~Wouter~Caarls

\thanks{This work has been submitted to the IEEE for possible publication. Copyright may be transferred without notice, after which this version may no longer be accessible.}
\thanks{This study was financed in part by the Coordenação de Aperfeiçoamento de Pessoal de Nível Superior - Brasil (CAPES) - Finance Code 001".}
\thanks{F. Cardeñoso and W. Caarls are with the Department
of Electrical Engineering, Pontificia Universidade Católica do Rio de Janeiro, Rio de Janeiro,
RJ, 22451-900 Brazil (e-mail: fracarfer5@aluno.puc-rio.br; wouter@puc-rio.br).}
}

%
%

\newcommand{\sortitem}[2]{%
	\DTLnewrow{list}%
	\DTLnewdbentry{list}{label}{#1}%
	\DTLnewdbentry{list}{description}{#2}%
}

\newenvironment{sortedlist}%
{%
	\DTLifdbexists{list}{\DTLcleardb{list}}{\DTLnewdb{list}}%
}%
{%
	\DTLsort{label}{list}%
	\begin{description}%
		\DTLforeach*{list}{\theLabel=label,\theDesc=description}{%
			\item[\theLabel] \theDesc
		}%
	\end{description}%
}

\newcommand{\tabitem}{~~\llap{\textbullet}~~}

\newcommand{\grad}{\nabla}
\newcommand{\vectorF}[1]{\pmb{#1}}
\newcommand{\dotVectorF}[1]{\pmb{\Dot{#1}}}
\newcommand{\vecF}[2]{\vectorF{#1}_{\mathrm{#2}}} 
\newcommand{\dotVecF}[2]{\dotVectorF{#1}_{\mathrm{#2}}}
\newcommand{\vecFqSd}{\vecF{q}{Sd}} 
\newcommand{\vecFqM}{\vecF{q}{M}}
\newcommand{\vecFqoffset}{\vecF{q}{offset}}
\newcommand{\vecFqMd}{\vecF{q}{Md}}
\newcommand{\vecFqS}{\vecF{q}{S}}
\newcommand{\vecFxSd}{\vecF{p}{Sd}}
\newcommand{\vecFxM}{\vecF{p}{M}}
\newcommand{\vecFxoffset}{\vecF{p}{offset}}
\newcommand{\vecFxMd}{\vecF{p}{Md}}
\newcommand{\vecFxS}{\vecF{p}{S}}
\newcommand{\scaleF}{\omega}
\newcommand{\scaleWSF}{\xi}
\newcommand{\datasetF}{\mathcal{D}}
\newcommand{\optPiF}{\pi^*}
\newcommand{\irlPiF}{\pi^\mathrm{E}}
\newcommand{\rewF}{r(\sF,\aF)} 
\newcommand{\returnF}{R_{t}} 
\newcommand{\rewardTF}[1]{r_{#1}} 
\newcommand{\expRewF}{\EF[ R_{t}]} 
\newcommand{\statesF}{\vectorF{s}_{t} \in S} 
\newcommand{\actionsF}{\vectorF{a}_{t} \in A}
\newcommand{\nxstatesF}{\vectorF{s}_{t+1} \in S}
\newcommand{\stateF}{\vectorF{s}_{t}}
\newcommand{\aF}{\vectorF{a}} 
\newcommand{\sF}{\vectorF{s}} 
\newcommand{\rF}{r} 
\newcommand{\nSF}{\vectorF{s}^{'}} 
\newcommand{\actionF}{\vectorF{a}_{t}}
\newcommand{\VpiF}{V^{\pi} (\sF)}
\newcommand{\nVpiF}{V^{\pi} (\sF_{t+1})}
\newcommand{\QpiF}{Q^{\pi} (\sF,\aF)}
\newcommand{\QpiFparam}{Q^{\pi_{\paramVF}} (\sF,\aF)}
\newcommand{\nQpiF}{Q^{\pi} (\sF_{t+1},\aF_{t+1})}
\newcommand{\EF}{\mathbb{E}}
\newcommand{\sumaF}[2]{\sum\limits ^{#2}_{i=#1}}
\newcommand{\optVpiF}{V^* (\sF)}
\newcommand{\optQpiF}{Q^* (\sF,\aF)}
\newcommand{\maxVF}{\underset{\pi}{\mathrm{max}}\ V^{\pi} (s)}
\newcommand{\maxF}[1]{\underset{#1}{\mathrm{max}}}
\newcommand{\maxQF}{\underset{\pi}{\mathrm{max}}\ Q^{\pi}(\sF,\aF)}
\newcommand{\paramVF}{\vectorF{\theta}} 
\newcommand{\paramVecF}{\vectorF{w}} 
\newcommand{\sDistF}{\rho^{\paramPolF}}
\newcommand{\paramPolF}{\pi_{\paramVF}}
\newcommand{\stoParamPolF}{\hat \pi(\aF \vert \sF;\actorPVF)} 
\newcommand{\stoParamVF}{\hat Q(\sF,\aF;\criticPVF)} 
\newcommand{\stateDistF}{\sF \sim \sDistF}
\newcommand{\actionDistF}{\aF \sim \paramPolF}
\newcommand{\detPolF}{\mu}
\newcommand{\detPQF}{Q^{\detPpolF}(\sF,\aF)} 
\newcommand{\detParamPolF}{\hat \detPolF(\sF;\actorPVF)} 
\newcommand{\ndetParamPolF}{\hat \detPolF(\sF';\actorPVF)} 
\newcommand{\idetParamPolF}{\hat \detPolF(\sF_{i};\actorPVF)} 
\newcommand{\detPpolF}{\detPolF_{\paramVF}}
\newcommand{\detParamQF}{\hat Q(\sF,\aF;\criticPVF)} 
\newcommand{\detParamAppQF}{\hat Q(\sF,\aF;\criticPVF)} 
\newcommand{\nDetParamAppQF}{\hat Q(\sF',\ntdetParamPolF;\targetCriticPVF)} 
\newcommand{\idetParamQF}{Q(\sF_{i}\idetParamPolF;\criticPVF)} 
\newcommand{\idetParamAppQF}{\hat Q(\sF_{i},\aF_{i};\criticPVF)} 
\newcommand{\tdetParamQF}{\hat Q(\sF,\aF;\targetCriticPVF)} 
\newcommand{\tdetParamPolF}{\hat \detPolF(\sF;\targetActorPVF)} 
\newcommand{\ndetParamQF}{\hat Q(\sF',\ntdetParamPolF;\targetCriticPVF)} 
\newcommand{\indetParamQF}{\hat Q(\sF'_i,\intdetParamPolF;\targetCriticPVF)} 
\newcommand{\ntdetParamPolF}{\hat \detPolF(\sF';\targetActorPVF)} 
\newcommand{\intdetParamPolF}{\hat \detPolF(\sF'_i;\targetActorPVF)} 
\newcommand{\iNdetParamQF}{\hat Q^{\detPolF}(\sF'_{i},\aF'_{i};\targetCriticPVF)} 
\newcommand{\actorPVF}{\paramVF} 
\newcommand{\criticPVF}{\paramVecF} 
\newcommand{\targetActorPVF}{\actorPVF^{\dagger}} 
\newcommand{\targetCriticPVF}{\criticPVF^{\dagger}} 
\newcommand{\ARP}{\mathcal{X}_{t}}
\newcommand{\arp}{AR-\textit{p}}
\newcommand{\ARPdisp}{\mathcal{X}_{t-i}}
\newcommand{\coeffF}{\Tilde{\phi}_{i}}
\newcommand{\gaussianF}{Z_{t}}
\newcommand{\desvF}{\tilde{\sigma}^{2}_{Z}}
\newcommand{\noiseF}{\mathcal{N} (0,\  \desvF)}
\newcommand{\detParamPolFtime}{\hat \detPolF (\sF_{t};\actorPVF)}
\newcommand{\noiseFtime}{\mathcal{N}_{t}}
\newcommand{\noiseProcess}{\mathcal{N}}
\newcommand{\arpProcess}{\mathcal{X}}
\newcommand{\shapeF}[3]{\mathrm{#1} \times \mathrm{#2} \times \mathrm{#3}}
\newcommand{\cnnShapeF}[1]{\Big[\frac{\mathrm{#1}+2p-k}{s}+1\Big]}
\newcommand{\positionVectorF}{\vectorF{p}} 
\newcommand{\velVectorF}{\dotVectorF{p}} 
\newcommand{\touchPosF}{\vecF{p}{H}}
\newcommand{\dobotPosF}{\vecF{p}{M}}
\newcommand{\dobotPosFd}{\vecF{p}{Md}}
\newcommand{\touchPosFscaled}{\vecF{\tilde{p}}{H}}
\newcommand{\dobotPosFscaled}{\vecF{\tilde{p}}{M}}
\newcommand{\taskF}{\mathit{task}}
\newcommand{\cVectorF}[1]{[x_{\mathrm{#1}},\ y_{\mathrm{#1}},\ z_{\mathrm{#1}}]}
\newcommand{\cVectorFscaled}[1]{\tilde{x}_{\mathrm{#1}},\ \tilde{y}_{\mathrm{#1}},\ \tilde{z}_{\mathrm{#1}}}
\newcommand{\VectorF}[1]{[\mathrm{#1}_{x},\ \mathrm{#1}_{y},\ \mathrm{#1}_{z}]}
\newcommand{\dotCVectorF}[1]{[\Dot{x}_{\mathrm{#1}},\ \Dot{y}_{\mathrm{#1}},\ \Dot{z}_{\mathrm{#1}}]}
\newcommand{\angleF}{\beta}
\newcommand{\finalRewF}{r_\mathrm{A}}
\newcommand{\forcesVectorF}{\vectorF{f}}
\newcommand{\fuzzyRewF}{r_\mathrm{F}}
\newcommand{\zeroRewF}{r_{0}}
\newcommand{\oneRewF}{r_{1}}
\newcommand{\fx}{f_{\mathrm{H} \parallel}}
\newcommand{\fy}{f_{\mathrm{H} \perp}}
\newcommand{\termF}{\varphi}
\newcommand{\normF}[1]{\Vert #1 \Vert}
\newcommand{\extraTermF}{x}
\newcommand{\dobotVelVF}{\velVectorF_{\mathrm{M}}}
\newcommand{\touchVelVF}{\velVectorF_{\mathrm{H}}}
\newcommand{\userForceF}{\forcesVectorF_{\mathrm{U}}}
\newcommand{\touchForceF}{\vecF{f}{H}}
\newcommand{\touchForceFmin}{\vecF{f}{H_{min}}}
\newcommand{\touchForceFmax}{\vecF{f}{H_{max}}}
\newcommand{\simPosVectorF}{\check{\vectorF{p}}}
\newcommand{\simTouchPosF}{\vectorF{\check{p}}_{\mathrm{H}}}
\newcommand{\simDobotPosF}{\vectorF{\check{p}}_{\mathrm{M}}}
\newcommand{\simVelVectorF}{\dotVectorF{\check{p}}}
\newcommand{\simUserForceF}{\vectorF{\check{f}}_{\mathrm{U}}}
\newcommand{\xF}[1]{x_\mathrm{#1}}
\newcommand{\yF}[1]{y_\mathrm{#1}}
\newcommand{\zF}[1]{z_\mathrm{#1}}
\newcommand{\discountF}{\gamma}
\newcommand{\finalStateF}{\sF_\mathrm{A}}
\newcommand{\bufferF}{\mathcal{R}} 
\newcommand{\batchF}{\mathcal{B}}
\newcommand{\absorbingF}{\mathrm{A}}
\newcommand{\statesSetF}{\mathcal{S}} 
\newcommand{\actionsSetF}{\mathcal{A}} 
\markboth{}%
{Cardeñoso and Caarls: Deep Reinforcement Learning for Haptic Shared Control in Unknown Tasks}

\maketitle

\begin{abstract}
Recent years have shown a growing interest in using haptic shared control (HSC) in teleoperated systems. In HSC, the application of virtual guiding forces decreases the user's control effort and improves execution time in various tasks, presenting a good alternative in comparison with direct teleoperation. HSC, despite demonstrating good performance, opens a new gap: how to design the guiding forces. For this reason, the challenge lies in developing controllers to provide the optimal guiding forces for the tasks that are being performed.
This work addresses this challenge by designing a controller based on the deep deterministic policy gradient (DDPG) algorithm to provide the assistance, and a convolutional neural network (CNN) to perform the task detection, called TAHSC (Task Agnostic Haptic Shared Controller). The agent learns to minimize the time it takes the human to execute the desired task, while simultaneously minimizing their resistance to the provided feedback. This resistance thus provides the learning algorithm with information about which direction the human is trying to follow, in this case, the \textit{pick-and-place} task.
Diverse results demonstrate the successful application of the proposed approach by learning custom policies for each user who was asked to test the system. It exhibits stable convergence and aids the user in completing the task with the least amount of time possible.
\end{abstract}


%
\IEEEpeerreviewmaketitle

\section{Introduction}
%
%
%
%

\IEEEPARstart{R}{obotic} systems exist everywhere. Because of their wide diversity, robots can be used for all kinds of applications. As a result, the coexistence between humans and robots has been growing in recent years and consequently, the necessity to develop human-robot collaboration to \textit{share control} to perform different tasks \cite{FLEMISCH201672}. A common example of this collaboration are teleoperated systems, where users control robots placed in remote locations. 
Thereby, the shared control (SC) approach becomes a useful tool combining the most powerful features of humans and robots for situations where humans cannot interact directly with the environment.
\begin{figure}
	\centering
	\includegraphics[width=.7\linewidth]{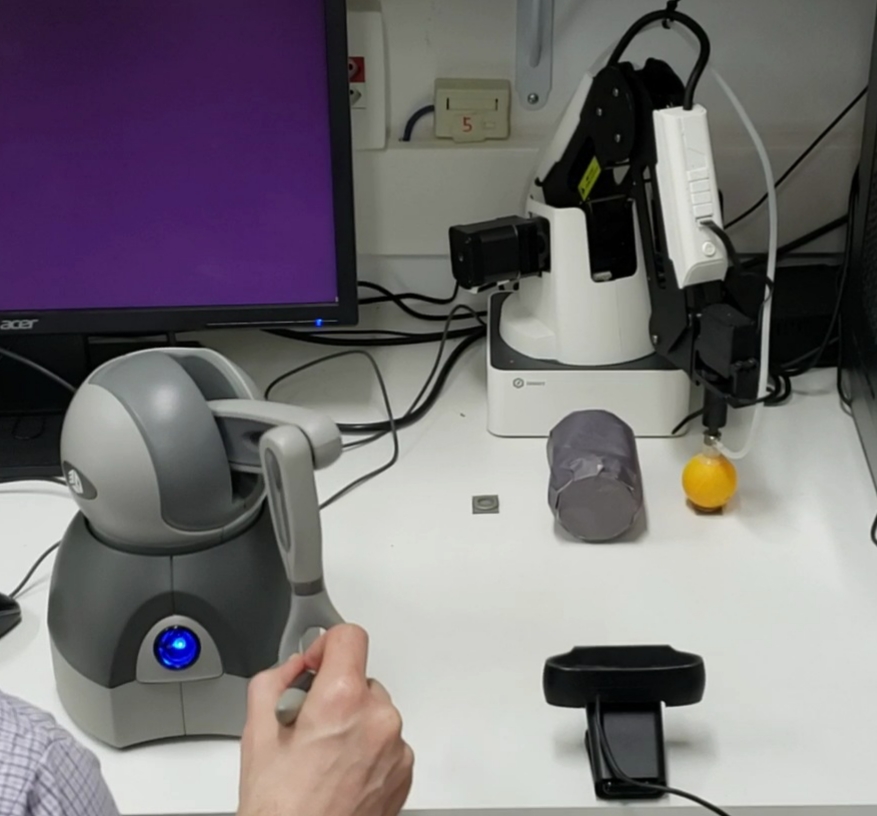}
	\caption{The user picks and places an object in a teleoperated system through a haptic device on the remote side and a robotic arm on the remote side.}
	\label{fig:test2}
\end{figure}
SC in teleoperation is not a new topic; earlier approaches demonstrated its performance and efficiency by applying optimal control and potential fields comparing direct and assistive control \cite{1043932}, learning how to modulate the control between the human and the robot semi autonomously~\cite{7989042} or learning how to produce forces to interact with unknown objects in virtual environments \cite{7759574}.
However, despite the fact that SC for teleoperation was widely addressed, optimal transparency has not been achieved and it has not yet been \textit{solved} in terms of optimizing the combined human-robot system's performance.

In this context, subsequent studies have shown that user performance is improved by \textit{decreasing} the transparency level and increasing the force feedback through the application of forces in the input device to guide the user movements. This decreases the user's control effort and improves the execution time to complete the task, obtaining in this way a new level of SC known as Haptic Shared Control (HSC) 
\cite{6212500}, \cite{7047839}. 
This kind of control, despite demonstrating good performance in teleoperated systems 
providing haptic guidance \cite{8794197}, \cite{7844727}, opens a new gap: how to design the guiding forces.

Moreover, although HSC has been sucessfully tested in a wide diversity of tasks, such as: peg-in-hole \cite{doi:10.1177/1729881419857428}, grasping objects \cite{7354028}, page turning task \cite{7333587}, home-service tasks \cite{7530842}, bolt-spanner task 
\cite{6212500}. It is possible to see that all of these controllers are designed to deal with fixed tasks or explicit information. In addition, it is known that real-world tasks are not always presented with fixed goals or trajectories. For this reason, the challenge turns on developing controllers to provide the guiding forces and at the same time able to deal with 
unknown tasks \cite{7247745}, \cite{6523033}.


Although linear and nonlinear control-based controllers present stable functionality in HSC for teleoperation as is described in \cite{8794197} or \cite{6907814}; it is necessary to have previous knowledge of the system model, dynamics and task.
In this context, the designed controllers should be able to handle unknown tasks as similar as humans do without much knowledge of the system. This kind of behavior is addressed by machine learning (ML) algorithms where performance is measured by the capability of abstraction and generalization.

In addition, to deal with situations where the task is unknown, it is desirable to implicitly give the controller some information about the task we are trying to do, so that the controller is able to compute the optimal behavior to complete the proposed task without explicit commands. Therefore, the algorithm only needs to be supplied with a learning function that reflects in a general way the user intention. This intention can be inferred from the user's actions, or implicitly by observing the environment and user behavior visually. Specifically, images can be passed to the learning controller giving implicit information about the task intention. This kind of information can be delivered to the controller directly or previously pre-processed with computer vision techniques. To achieve generalization in the image processing, Convolutional Neural Networks (CNN) are a good option, to provide the controller with the relevant features of the task.

Therefore, given that the real challenge remains in designing a controller with enough adaptability to assist the user it is necessary to choose an approach able to provide this feature. Reinforcement Learning (RL)~\cite{sutton} is an interesting option because it brings us the possibility to \textit{teach} the controller what our intention is through trial and error interaction. The general learning rules are encoded in the \textit{reward function} which is used as a metric to measure the performance of the controller during the training. Thus, learning the system model is not necessary (such as in control engineering techniques) and we only need to supply the algorithm with the observations composed of relevant information about the current states every time step the algorithm is executed. Moreover, the information of the task intention can be provided using a camera, sending visual information which is then processed by the designed controller, thereby taking as inputs information composed by measurements and images, and giving as outputs the needed assistance with the guide forces to complete the task. 
The use of RL in HSC has been previously tested with interesting results as present~\cite{Ewertonetal18},
\cite{Ewerton_IROS_2019} and \cite{5975338}.

In order to address this challenge, this work aims to develop an HSC controller based on RL to learn the assistance forces and performs the task detection autonomously with a camera on a teleoperated system as presents the Figure~\ref{fig:test2}. Experiments with different subjects were carried out to validate the implemented controller. In addition, to perform the preliminary tests, the implementation of a simulator is also proposed. Unlike previous studies, as the main contribution, our approach provides the assistance forces and decodes the task intention without prior knowledge of the task. Thus, our system learns online and with a limited number of epochs from samples of successful and unsuccessful demonstrations using numerical and visual information. As a secondary contribution, we present the implementation of a simulator that can be used to analyze the system performance. Then, different modifications and hyperparameter settings can be tested before being used in the real system.
\\

\section{System Description}\label{section:systemDescription}

\begin{figure*}
	\centering 
	\resizebox{\linewidth}{!}{\input{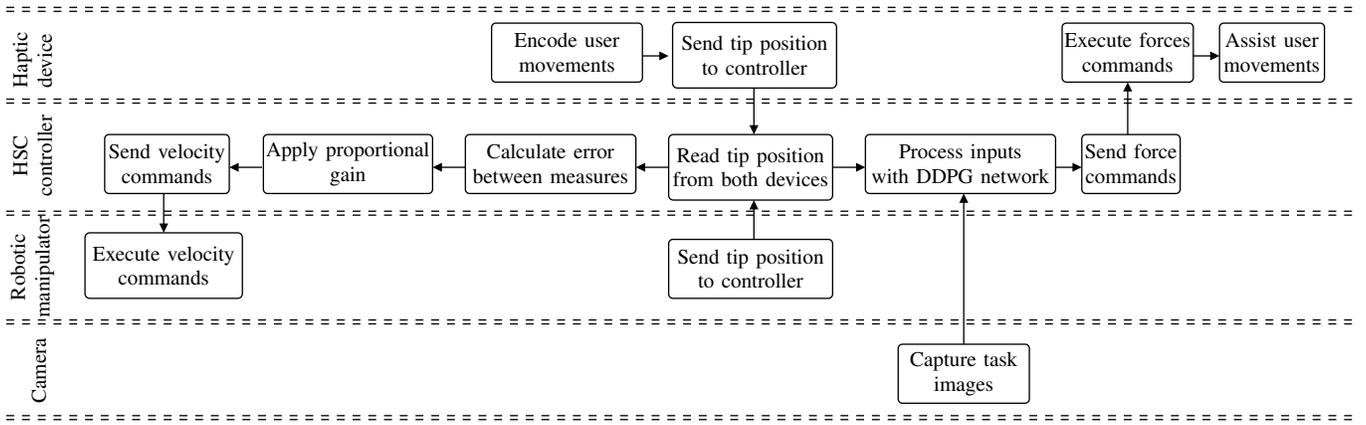}}
	\caption{HSC controller flowchart.}
	\label{fig:teleoperationFlowchart}
\end{figure*}

As was discussed in the previous section, HSC has shown promising results improving the operator's performance in teleoperation \cite{6212500}, \cite{7047839}. However, practically all applications that use this approach learn at their core of a set of trajectory distributions \cite{7989042}, \cite{Ewerton_IROS_2019}. Although it is well known the importance of inferring user's goal from his actions in shared autonomy, recent research still uses fixed goals \cite{doi:10.1177/0278364918776060}.

Therefore, despite the good performance achieved with these algorithms, we can not consider them if the intention is to help the user to perform tasks that depend on implicit information, for example current images of the performed task. We propose to address the problem of implicitly encoding task intention using two types of data: visual and numerical. In this way, it is possible to perform any task that is able to be encoded through the state vector.



The proposed system bases its operation in the central controller which is composed of two separate sub-controllers: the \textit{teleoperation controller} and  the \textit{RL-controller}. The teleoperation controller is responsible for \textit{replicating} the master movements on the slave side, while the RL-controller is responsible for processing the information and sending the commands to the haptic device in order to assist the operator. The basic functioning of the two controllers can be explained by the flowcharts presented in Figure~\ref{fig:teleoperationFlowchart}.

The chosen task will be the \textit{pick-and-place} task as it is a very common task in domestic and industrial applications. This task consists of picking an object from a \textit{initial point}, and dropping it in a \textit{goal} position. 
To increase complexity in performing the task, an obstacle will be placed between the initial and the final points as shown in Figure~\ref{fig:test2}. In addition, the pick-and-place task will be performed in a bidirectional way; that is, take the ball from the initial point to the goal, and then, in the next half, the point positions are exchanged; this modification was added to demonstrate that the proposed system is able to learn more than one task. So, a complete epoch is considered to following the next trajectory: \textit{initial point-goal-initial point}.


\section{TAHSC controller}\label{section:method}
As was mentioned in Section \ref{section:systemDescription}, the proposed HSC controller is composed of two sub-controllers: the teleoperation and the RL-controller. In this section, we describe their functioning. 
\subsection{Teleoperation controller}\label{section:teleoperation controller}
The \textit{teleoperation controller} is responsible for replicating the user movements with the Haptic device in the manipulator. This is performed through direct teleoperation with position control in the tip level using the position coordinates of the haptic device $\touchPosF$ and the manipulator $\dobotPosF$ respectively\cite{DBLP:reference/robo/NiemeyerPSL16}.


The direct teleoperation is achieved in the following way: Let $\touchPosF=\cVectorF{H}$ and $\dobotPosF=\cVectorF{M}$ be the current position of the end-effector for the haptic device and the manipulator with respect to their base link in Cartesian coordinates respectively.
The error between the desired and the current position is calculated by:
\begin{align}
\vectorF{e} &= \dobotPosFd - \dobotPosF \,,\\
\dobotPosFd &= \scaleWSF \touchPosF + \vecFxoffset \,.
\label{eq:errorVector}
\end{align}
where $\vectorF{e} = \VectorF{e}$ are the calculated errors for axis $x, y$ and $z$ respectively and $\scaleWSF$ is a scale factor. The manipulator velocity to track the haptic device tip is calculated by multiplying a proportional positive gain $k_{M}$ to the error vector calculated in~(\ref{eq:errorVector}):
\begin{equation}
\dotVecF{p}{M} = k_{M} \vectorF{e} \,.
\label{eq:velocityVector}
\end{equation}
where $\dobotVelVF = \dotCVectorF{M}$ are the calculated velocities for the manipulator end-effector. Thereby, the manipulator replicates the haptic device movements with a velocity proportional to the error between both devices.

\subsection{RL-controller}

The RL-controller is responsible for performing the \textit{assistance} to the teleoperator. This controller bases its operation in the DDPG algorithm explained in \cite{DBLP:journals/corr/LillicrapHPHETS15}. To perform the assistance, the DDPG network is trained with relevant information about the positions as inputs and providing the guiding forces as outputs. 

\subsubsection{DDPG state vector}\label{section:DDPGstatesVector}
Handles the information coming from both devices and the task detector. This state vector is used in the DDPG network and stored in the replay buffer with other samples to train the agent. The proposed state vector for the DDPG network consists of the tip positions $\dobotPosF$ and $\touchPosF$ respectively. In addition, an extra term called $\mathit{task}$ is appended which encodes the task intention: 
\begin{equation}
\sF = [\dobotPosF,\  \touchPosF,\  \taskF] \,,
\label{eq:stateVector}
\end{equation}
Moreover, it is common that the input vector for ANNs is in the range of $[-1,\ 1]$, therefore, the position vectors are scaled in an accepted range before being processed by the DDPG network using a scale factor $\scaleF$.

This value was chosen as the maximum value between the largest possible value that both devices can reach in their respective workspaces:
\begin{equation}
\scaleF = \max (\max (\dobotPosF), \max (\touchPosF))\,.
\label{eq:scaleFactor}
\end{equation}

In this way, the result of applying a scaling in the state vector given in~(\ref{eq:stateVector}) is:
\begin{equation}
\sF = [\dobotPosFscaled,\ \touchPosFscaled,\ \taskF]\,,
\label{eq:stateVectorscaled}
\end{equation} 
where $\dobotPosFscaled=\frac{\dobotPosF}{\scaleF}$ and  $\touchPosFscaled=\frac{\touchPosF}{\scaleF}$. 

\subsubsection{DDPG network architecture}\label{section:DDPGnetworkArchitecture}

The network architecture implemented for this approach was adapted from the original configuration described in~\cite{DBLP:journals/corr/LillicrapHPHETS15}, where the actor network consists of an ANN with three layers of 400, 300 and 3 units with \textit{ReLU} activation for the first two layers and \textit{tanh} activation in the last layer. Moreover, the input size is the states vectors $\sF$ described in subsection~\ref{section:DDPGstatesVector} and the output is the assistive force vector $\touchForceF$.

The critic network is composed of three layers of 400, 300 and 1 units with \textit{ReLU} activation for the first two layers and \textit{linear} activation in the last layer respectively. The input size is similar to the actor network with the difference that the action vector $\aF$ is concatenated in the second hidden layer. The critic network output is the Q-value for the given input state vectors. 


The replay buffer $\bufferF$ is a set of tuples, where each tuple is composed of the previous observation $\sF$, the action $\aF$, the reward $r$, the current observation $\sF'$ and an extra element termed $\absorbingF$ which indicates if the state was terminal or not: 
\begin{equation}
\bufferF \leftarrow  \bufferF \cup \{(\sF,\aF,r,\sF', \absorbingF)\}
\end{equation}

It is worth noticing that~(\ref{eq:stateVector}) presents an extra term in the state vector that is appended to encode the task information. To perform this \textit{task coding} in the DDPG network, we consider a new functionality in the proposed system called the \textit{task detector}. 

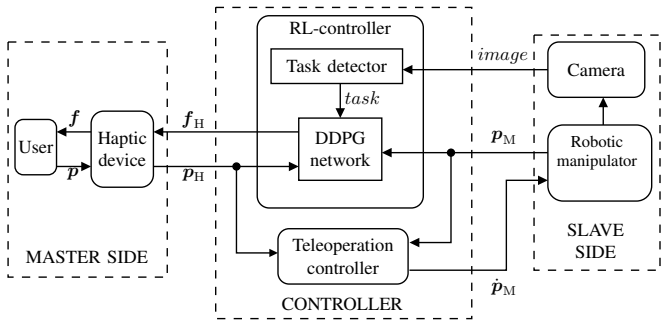
\begin{figure}
	\centering 
	\resizebox{\linewidth}{!}{\tikzset{every picture/.style={line width=0.75pt}} 

\begin{tikzpicture}[x=0.75pt,y=0.75pt,yscale=-1,xscale=1]

\draw    (265,4365) -- (367,4365) ;
\draw [shift={(370,4365)}, rotate = 180] [fill={rgb, 255:red, 0; green, 0; blue, 0 }  ][line width=0.08]  [draw opacity=0] (7.14,-3.43) -- (0,0) -- (7.14,3.43) -- cycle    ;
\draw [line width=0.75]    (450,4440) -- (520,4440) -- (520,4375) -- (547,4375) ;
\draw [shift={(550,4375)}, rotate = 180] [fill={rgb, 255:red, 0; green, 0; blue, 0 }  ][line width=0.08]  [draw opacity=0] (7.14,-3.43) -- (0,0) -- (7.14,3.43) -- cycle    ;
\draw    (550,4355) -- (433,4355) ;
\draw [shift={(430,4355)}, rotate = 360] [fill={rgb, 255:red, 0; green, 0; blue, 0 }  ][line width=0.08]  [draw opacity=0] (7.14,-3.43) -- (0,0) -- (7.14,3.43) -- cycle    ;
\draw    (480,4355) -- (480,4420) -- (453,4420) ;
\draw [shift={(450,4420)}, rotate = 360] [fill={rgb, 255:red, 0; green, 0; blue, 0 }  ][line width=0.08]  [draw opacity=0] (7.14,-3.43) -- (0,0) -- (7.14,3.43) -- cycle    ;
\draw  [fill={rgb, 255:red, 0; green, 0; blue, 0 }  ,fill opacity=1 ] (322.5,4365) .. controls (322.5,4363.62) and (323.62,4362.5) .. (325,4362.5) .. controls (326.38,4362.5) and (327.5,4363.62) .. (327.5,4365) .. controls (327.5,4366.38) and (326.38,4367.5) .. (325,4367.5) .. controls (323.62,4367.5) and (322.5,4366.38) .. (322.5,4365) -- cycle ;
\draw   (340,4264.62) .. controls (340,4259.86) and (343.86,4256) .. (348.62,4256) -- (451.38,4256) .. controls (456.14,4256) and (460,4259.86) .. (460,4264.62) -- (460,4386.38) .. controls (460,4391.14) and (456.14,4395) .. (451.38,4395) -- (348.62,4395) .. controls (343.86,4395) and (340,4391.14) .. (340,4386.38) -- cycle ;
\draw    (400,4327) -- (400,4305) ;
\draw [shift={(400,4330)}, rotate = 270] [fill={rgb, 255:red, 0; green, 0; blue, 0 }  ][line width=0.08]  [draw opacity=0] (7.14,-3.43) -- (0,0) -- (7.14,3.43) -- cycle    ;
\draw  [fill={rgb, 255:red, 0; green, 0; blue, 0 }  ,fill opacity=1 ] (477.5,4355) .. controls (477.5,4353.62) and (478.62,4352.5) .. (480,4352.5) .. controls (481.38,4352.5) and (482.5,4353.62) .. (482.5,4355) .. controls (482.5,4356.38) and (481.38,4357.5) .. (480,4357.5) .. controls (478.62,4357.5) and (477.5,4356.38) .. (477.5,4355) -- cycle ;
\draw    (195,4365) -- (217,4365) ;
\draw [shift={(220,4365)}, rotate = 180] [fill={rgb, 255:red, 0; green, 0; blue, 0 }  ][line width=0.08]  [draw opacity=0] (7.14,-3.43) -- (0,0) -- (7.14,3.43) -- cycle    ;
\draw    (220,4340) -- (198,4340) ;
\draw [shift={(195,4340)}, rotate = 360] [fill={rgb, 255:red, 0; green, 0; blue, 0 }  ][line width=0.08]  [draw opacity=0] (7.14,-3.43) -- (0,0) -- (7.14,3.43) -- cycle    ;
\draw   (165,4337) .. controls (165,4333.69) and (167.69,4331) .. (171,4331) -- (189,4331) .. controls (192.31,4331) and (195,4333.69) .. (195,4337) -- (195,4365) .. controls (195,4368.31) and (192.31,4371) .. (189,4371) -- (171,4371) .. controls (167.69,4371) and (165,4368.31) .. (165,4365) -- cycle ;
\draw   (220,4334) .. controls (220,4329.03) and (224.03,4325) .. (229,4325) -- (256,4325) .. controls (260.97,4325) and (265,4329.03) .. (265,4334) -- (265,4371) .. controls (265,4375.97) and (260.97,4380) .. (256,4380) -- (229,4380) .. controls (224.03,4380) and (220,4375.97) .. (220,4371) -- cycle ;
\draw  [dash pattern={on 4.5pt off 4.5pt}] (160,4275) -- (275,4275) -- (275,4445) -- (160,4445) -- cycle ;
\draw   (550,4346) .. controls (550,4339.92) and (554.92,4335) .. (561,4335) -- (614,4335) .. controls (620.08,4335) and (625,4339.92) .. (625,4346) -- (625,4379) .. controls (625,4385.08) and (620.08,4390) .. (614,4390) -- (561,4390) .. controls (554.92,4390) and (550,4385.08) .. (550,4379) -- cycle ;
\draw  [dash pattern={on 4.5pt off 4.5pt}] (540,4270) -- (631,4270) -- (631,4440) -- (540,4440) -- cycle ;
\draw   (550,4283) .. controls (550,4278.58) and (553.58,4275) .. (558,4275) -- (612,4275) .. controls (616.42,4275) and (620,4278.58) .. (620,4283) -- (620,4307) .. controls (620,4311.42) and (616.42,4315) .. (612,4315) -- (558,4315) .. controls (553.58,4315) and (550,4311.42) .. (550,4307) -- cycle ;
\draw    (590,4335) -- (590,4318) ;
\draw [shift={(590,4315)}, rotate = 450] [fill={rgb, 255:red, 0; green, 0; blue, 0 }  ][line width=0.08]  [draw opacity=0] (7.14,-3.43) -- (0,0) -- (7.14,3.43) -- cycle    ;
\draw    (370,4340) -- (268,4340) ;
\draw [shift={(265,4340)}, rotate = 360] [fill={rgb, 255:red, 0; green, 0; blue, 0 }  ][line width=0.08]  [draw opacity=0] (7.14,-3.43) -- (0,0) -- (7.14,3.43) -- cycle    ;
\draw    (325,4365) -- (325,4427.5) -- (352,4427.5) ;
\draw [shift={(355,4427.5)}, rotate = 180] [fill={rgb, 255:red, 0; green, 0; blue, 0 }  ][line width=0.08]  [draw opacity=0] (6.25,-3) -- (0,0) -- (6.25,3) -- cycle    ;
\draw   (355,4418) .. controls (355,4413.58) and (358.58,4410) .. (363,4410) -- (441,4410) .. controls (445.42,4410) and (449,4413.58) .. (449,4418) -- (449,4442) .. controls (449,4446.42) and (445.42,4450) .. (441,4450) -- (363,4450) .. controls (358.58,4450) and (355,4446.42) .. (355,4442) -- cycle ;
\draw   (370,4330) -- (430,4330) -- (430,4375) -- (370,4375) -- cycle ;
\draw   (350,4280) -- (445,4280) -- (445,4305) -- (350,4305) -- cycle ;
\draw    (550,4295) -- (448,4295) ;
\draw [shift={(445,4295)}, rotate = 360] [fill={rgb, 255:red, 0; green, 0; blue, 0 }  ][line width=0.08]  [draw opacity=0] (7.14,-3.43) -- (0,0) -- (7.14,3.43) -- cycle    ;
\draw  [dash pattern={on 4.5pt off 4.5pt}] (310,4250) -- (495,4250) -- (495,4475) -- (310,4475) -- cycle ;

\draw (517.5,4285) node    {$image$};
\draw (518.5,4344) node    {$\boldsymbol{p}_{\mathrm{M}}$};
\draw (401.5,4464.5) node   [align=left] {CONTROLLER};
\draw (518.5,4452) node    {$\dot{\boldsymbol{p}}_{\mathrm{M}}$};
\draw (400,4264.5) node   [align=left] {RL-controller};
\draw (416,4315) node    {$task$};
\draw (400,4352.5) node   [align=left] {\begin{minipage}[lt]{38.454pt}\setlength\topsep{0pt}
\begin{center}
DDPG\\network
\end{center}

\end{minipage}};
\draw (402,4430) node   [align=left] {\begin{minipage}[lt]{64.01064400000001pt}\setlength\topsep{0pt}
\begin{center}
Teleoperation\\controller
\end{center}

\end{minipage}};
\draw (217,4430.5) node   [align=left] {MASTER SIDE};
\draw (207,4370) node    {$\boldsymbol{p}$};
\draw (208,4330) node    {$\boldsymbol{f}$};
\draw (180,4351) node   [align=left] {User};
\draw (242.5,4352.5) node   [align=left] {\begin{minipage}[lt]{32.220644pt}\setlength\topsep{0pt}
\begin{center}
Haptic\\device
\end{center}

\end{minipage}};
\draw (584.5,4419) node   [align=left] {\begin{minipage}[lt]{34.838644pt}\setlength\topsep{0pt}
\begin{center}
SLAVE\\SIDE
\end{center}

\end{minipage}};
\draw (585.5,4355) node  [font=\small] [align=left] {\begin{minipage}[lt]{50.660000000000004pt}\setlength\topsep{0pt}
\begin{center}
Robotic\\manipulator
\end{center}

\end{minipage}};
\draw (585,4295) node   [align=left] {Camera};
\draw (294.5,4371) node    {$\boldsymbol{p}_{\mathrm{H}}$};
\draw (294,4331) node    {$\boldsymbol{f}_{\mathrm{H}}$};
\draw (397.5,4292.5) node   [align=left] {Task detector};

\end{tikzpicture}}
	\caption{HSC controller with camera input functionality.}
	\label{fig:teleoperation camera input}
\end{figure}

We assume the camera images contain sufficient information to identify the task, which is decoded using a CNN. Thereby, the operator does not have to provide any extra information while is performing the task. Then the system can detect the task autonomously, as well as learn the task detection on-the-fly. Figure~\ref{fig:teleoperation camera input} summarizes the proposed implementation.

So, in order to perform the task detection, a set of modifications was made in the original DDPG implementation. First of all, we apply the 
pre-trained CNN VGG16~\cite{Simonyan15} to extract the visual features of the task, and then, the resulting output is passed into the states vector. Thereby, the DDPG agent learns from visual information decoding which is the task that is being performed. In addition, since we are using RGB images, it is important to take into account that storing visual information in the replay buffer ($\bufferF$) can lead to a memory saturation of the computer. Moreover, the training time used increases due to the amount of parameters of the complete network (VGG16 + DDPG). To deal with this situation, we propose other modification: the resulting VGG16 network is separated in two parts, from the input until the \textit{GlobalAveragePooling2D} layer, which is used as frozen model, and the second part, which consist in a dense layer that is appended with the DDPG network. In this way, the first separated part of the CNN network is used to provide the resulting features of the images and the second part is trained with the DDPG network. Finally, to decrease the inference time, the initial frame, which is taken in the first time step, is passed through the CNN network and the resulting output values are maintained along the entire episode. Using this modification the number of parameters to be trained and the inference time are decreased resulting in less time to train the HSC controller. 

Note that this approach will only be effective if the addressed task can be determined with visual information of the initial position, for example the pick-and-place task in this case. However, this information will not sufficient in for example the peg-in-hole task. In this case, both, the first and the current image features should be added to the states vector, thereby allowing virtually any task intention to be decoded.

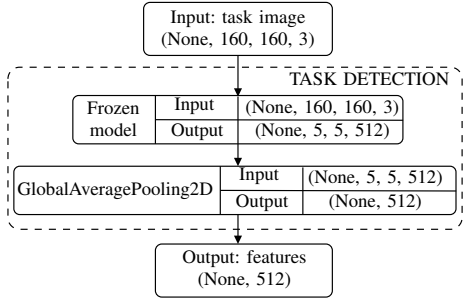
\begin{figure}
	\centering 
	\resizebox{0.7\linewidth}{!}{\tikzset{every picture/.style={line width=0.75pt}} 

\begin{tikzpicture}[x=0.75pt,y=0.75pt,yscale=-1,xscale=1]

\draw  [dash pattern={on 4.5pt off 4.5pt}] (120,2119.87) .. controls (120,2114.42) and (124.42,2110) .. (129.87,2110) -- (495.13,2110) .. controls (500.58,2110) and (505,2114.42) .. (505,2119.87) -- (505,2237.63) .. controls (505,2243.08) and (500.58,2247.5) .. (495.13,2247.5) -- (129.87,2247.5) .. controls (124.42,2247.5) and (120,2243.08) .. (120,2237.63) -- cycle ;
\draw   (245,2265.46) .. controls (245,2262.45) and (247.45,2260) .. (250.46,2260) -- (389.54,2260) .. controls (392.55,2260) and (395,2262.45) .. (395,2265.46) -- (395,2299.54) .. controls (395,2302.55) and (392.55,2305) .. (389.54,2305) -- (250.46,2305) .. controls (247.45,2305) and (245,2302.55) .. (245,2299.54) -- cycle ;
\draw    (315,2235) -- (315,2257) ;
\draw [shift={(315,2260)}, rotate = 270] [fill={rgb, 255:red, 0; green, 0; blue, 0 }  ][line width=0.08]  [draw opacity=0] (7.14,-3.43) -- (0,0) -- (7.14,3.43) -- cycle    ;
\draw    (300,2194) -- (300,2235) ;
\draw    (300,2214.5) -- (500,2215) ;
\draw    (315,2175) -- (315,2191) ;
\draw [shift={(315,2194)}, rotate = 270] [fill={rgb, 255:red, 0; green, 0; blue, 0 }  ][line width=0.08]  [draw opacity=0] (7.14,-3.43) -- (0,0) -- (7.14,3.43) -- cycle    ;
\draw   (125,2198.98) .. controls (125,2196.23) and (127.23,2194) .. (129.98,2194) -- (494.84,2194) .. controls (497.59,2194) and (499.82,2196.23) .. (499.82,2198.98) -- (499.82,2230.02) .. controls (499.82,2232.77) and (497.59,2235) .. (494.84,2235) -- (129.98,2235) .. controls (127.23,2235) and (125,2232.77) .. (125,2230.02) -- cycle ;
\draw    (370,2195) -- (370,2235) ;
\draw    (245,2134) -- (245,2175) ;
\draw    (245,2154.5) -- (455,2155) ;
\draw   (175,2138.98) .. controls (175,2136.23) and (177.23,2134) .. (179.98,2134) -- (449.84,2134) .. controls (452.59,2134) and (454.82,2136.23) .. (454.82,2138.98) -- (454.82,2170.02) .. controls (454.82,2172.77) and (452.59,2175) .. (449.84,2175) -- (179.98,2175) .. controls (177.23,2175) and (175,2172.77) .. (175,2170.02) -- cycle ;
\draw    (315,2135) -- (315,2175) ;
\draw    (315,2100) -- (315,2132) ;
\draw [shift={(315,2135)}, rotate = 270] [fill={rgb, 255:red, 0; green, 0; blue, 0 }  ][line width=0.08]  [draw opacity=0] (7.14,-3.43) -- (0,0) -- (7.14,3.43) -- cycle    ;
\draw   (235,2060.46) .. controls (235,2057.45) and (237.45,2055) .. (240.46,2055) -- (389.54,2055) .. controls (392.55,2055) and (395,2057.45) .. (395,2060.46) -- (395,2094.54) .. controls (395,2097.55) and (392.55,2100) .. (389.54,2100) -- (240.46,2100) .. controls (237.45,2100) and (235,2097.55) .. (235,2094.54) -- cycle ;

\draw (431.5,2204.5) node  [font=\large] [align=left] {(None, 5, 5, 512)};
\draw (431.5,2224.5) node  [font=\large] [align=left] {(None, 512)};
\draw (211,2214.5) node  [font=\large] [align=left] {GlobalAveragePooling2D};
\draw (335.5,2224.5) node  [font=\large] [align=left] {Output};
\draw (333,2202.5) node  [font=\large] [align=left] {Input};
\draw (386.5,2144.5) node  [font=\large] [align=left] {(None, 160, 160, 3)};
\draw (386.5,2164.5) node  [font=\large] [align=left] {(None, 5, 5, 512)};
\draw (209.5,2154) node  [font=\large] [align=left] {\begin{minipage}[lt]{40.834pt}\setlength\topsep{0pt}
\begin{center}
Frozen\\model
\end{center}

\end{minipage}};
\draw (280.5,2164.5) node  [font=\large] [align=left] {Output};
\draw (278,2142.5) node  [font=\large] [align=left] {Input};
\draw (315,2077.5) node  [font=\large] [align=left] {\begin{minipage}[lt]{108.23356000000001pt}\setlength\topsep{0pt}
\begin{center}
Input: task image\\(None, 160, 160, 3)
\end{center}

\end{minipage}};
\draw (426.5,2119.5) node  [font=\large] [align=left] {TASK DETECTION};
\draw (320.5,2281) node  [font=\large] [align=left] {\begin{minipage}[lt]{114.34200000000001pt}\setlength\topsep{0pt}
\begin{center}
Output: features\\(None, 512)
\end{center}

\end{minipage}};

\end{tikzpicture}}
	\caption{Task detector CNN network architecture used.}
	\label{fig:ddpg2 task detector CNN} 
\end{figure}

\begin{figure}
	\centering 
	\resizebox{\linewidth}{!}{\input{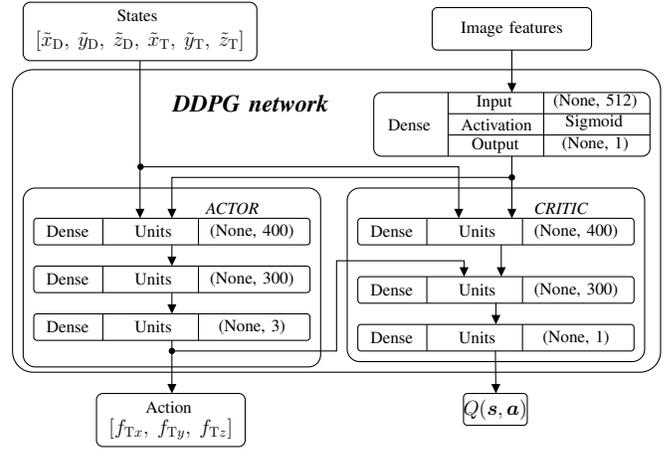}}
	\caption{RL-controller architecture for the TAHSC approach implemented.}
	\label{fig:ddpg2 second approach} 
\end{figure}

Figures~\ref{fig:ddpg2 task detector CNN} and~\ref{fig:ddpg2 second approach} present the mounted network with the proposed modifications. Notice that the VGG16 network is separated and is not trained within the DDPG network. In this way, only the output of the dense layer is concatenated with the position vectors. 
We named this approach as Task Agnostic Haptic Shared Control (TAHSC), or simply TAHSC approach. Algorithm~\ref{alg:ddpg 2_2} shows the implementation of the TAHSC approach.



\subsubsection{Reward function design}
As explained in \cite{DBLP:reference/robo/0001LKNBS16}, one of the underestimated problems in RL is the goal specification, which is achieved by designing a proper reward function. This function must capture the general learning rules allowing the RL algorithm to converge to the desired behavior. 

The reward function used for our approach bases its operation on the difference of magnitude between the assistive force $\touchForceF$ and the user velocity $\touchVelVF$ presented in Figure~\ref{fig:reward2}, combined with fuzzy rules that enhance the functioning of the reward function in order to capture more of the user intention and boost the agent learning. Thereby, this function is designed to encourage the agent to influence the user to execute the task quickly and give him more control in the task execution; to achieve that, fuzzy rules condition the amount of reward that the agent receives according the velocity of the user.

\begin{figure}
	\centering
	\resizebox{0.5\linewidth}{!}{\tikzset{every picture/.style={line width=0.75pt}} 

\begin{tikzpicture}[x=0.75pt,y=0.75pt,yscale=-1,xscale=1]

\draw [line width=1.5]    (566.24,3076.82) .. controls (567.74,3066.32) and (565.24,3058.32) .. (560.61,3051.34) ;
\draw [color={rgb, 255:red, 0; green, 0; blue, 0 }  ,draw opacity=1 ][fill={rgb, 255:red, 208; green, 2; blue, 27 }  ,fill opacity=1 ][line width=1.5]    (522.63,3076.58) -- (682.32,2970.5) ;
\draw [shift={(685.65,2968.29)}, rotate = 506.41] [fill={rgb, 255:red, 0; green, 0; blue, 0 }  ,fill opacity=1 ][line width=0.08]  [draw opacity=0] (11.61,-5.58) -- (0,0) -- (11.61,5.58) -- cycle    ;
\draw [color={rgb, 255:red, 0; green, 0; blue, 0 }  ,draw opacity=1 ][line width=1.5]  [dash pattern={on 5.63pt off 4.5pt}]  (685.65,2968.29) -- (685.65,3078.42) ;
\draw [color={rgb, 255:red, 0; green, 0; blue, 0 }  ,draw opacity=1 ][line width=1.5]  [dash pattern={on 5.63pt off 4.5pt}]  (522.63,2966.45) -- (522.63,3076.58) ;
\draw [color={rgb, 255:red, 0; green, 0; blue, 0 }  ,draw opacity=1 ][line width=1.5]  [dash pattern={on 5.63pt off 4.5pt}]  (522.63,2966.45) -- (685.65,2968.29) ;
\draw [color={rgb, 255:red, 0; green, 0; blue, 0 }  ,draw opacity=1 ][line width=1.5]    (522.63,3076.58) -- (768.63,3076.58) ;
\draw [shift={(772.63,3076.58)}, rotate = 180] [fill={rgb, 255:red, 0; green, 0; blue, 0 }  ,fill opacity=1 ][line width=0.08]  [draw opacity=0] (11.61,-5.58) -- (0,0) -- (11.61,5.58) -- cycle    ;

\draw (608,2954) node  [font=\Large,color={rgb, 255:red, 0; green, 0; blue, 0 }  ,opacity=1 ]  {$f_{\mathrm{H} \parallel } =\| \boldsymbol{f}_{\mathrm{H}} \| \cos \beta $};
\draw (577.5,3058) node  [font=\Large]  {$\beta $};
\draw (762.5,3021) node  [font=\Large,color={rgb, 255:red, 0; green, 0; blue, 0 }  ,opacity=1 ]  {$f_{\mathrm{H} \perp } =\| \boldsymbol{f}_{\mathrm{H}} \| \sin \beta $};
\draw (593.54,3010.22) node  [font=\Large,color={rgb, 255:red, 0; green, 0; blue, 0 }  ,opacity=1 ,rotate=-329.62]  {$\boldsymbol{f}_{\mathrm{H}}$};
\draw (638,3092.5) node  [font=\LARGE,color={rgb, 255:red, 0; green, 0; blue, 0 }  ,opacity=1 ]  {$\dot{\boldsymbol{p}_{\mathrm{H}}}$};

\end{tikzpicture}}
	\caption{Reward function diagram: calculates the rewards based in the magnitude of the $\touchForceF$ and $\touchVelVF$ vectors.}
	\label{fig:reward2}
\end{figure}
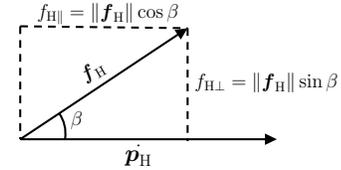
\begin{figure}
	\centering
	\resizebox{0.65\linewidth}{!}{\tikzset{every picture/.style={line width=0.75pt}} 

\begin{tikzpicture}[x=0.75pt,y=0.75pt,yscale=-1,xscale=1]

\draw [line width=1.5]    (135,3189.95) -- (249.37,3189.95) ;
\draw [line width=1.5]    (135,3275) -- (249.37,3275) ;
\draw [line width=1.5]    (322.68,3189.95) -- (522.4,3189.95) ;
\draw [line width=1.5]    (322.68,3275) -- (526,3274.67) ;
\draw [shift={(530,3274.66)}, rotate = 539.9100000000001] [fill={rgb, 255:red, 0; green, 0; blue, 0 }  ][line width=0.08]  [draw opacity=0] (11.61,-5.58) -- (0,0) -- (11.61,5.58) -- cycle    ;
\draw [line width=1.5]  [dash pattern={on 5.63pt off 4.5pt}]  (249.37,3275) -- (322.68,3275) ;
\draw [line width=1.5]    (249.37,3189.95) -- (322.68,3275) ;
\draw [line width=1.5]    (249.37,3275) -- (322.68,3189.95) ;

\draw (193,3175.5) node  [font=\LARGE] [align=left] {$\displaystyle r_{0} =-\| \boldsymbol{f}_{\mathrm{H}}\Vert $};
\draw (420.5,3175.5) node  [font=\LARGE] [align=left] {$\displaystyle r_{1} =-f_{\mathrm{H} \perp } +f_{\mathrm{H} \parallel } +c$};
\draw (549,3294) node  [font=\LARGE] [align=left] {$\displaystyle \| \boldsymbol{\dot{p}}_{\mathrm{H}}\Vert $};
\draw (250.5,3286) node  [font=\LARGE] [align=left] {$\displaystyle 0$};
\draw (320.5,3286) node  [font=\LARGE] [align=left] {$\displaystyle x$};

\end{tikzpicture}}
	\caption{Reward function: conditional rules.}
	\label{fig:fuzzyReward2}
\end{figure}
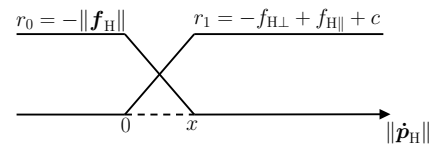
The functioning of the designed reward function is described in the following:
\begin{align}
\rewF &=\begin{cases}
\finalRewF & ,\ \text{if $\sF$ is an absorbing state}\\
\fuzzyRewF\  & ,\ \text{otherwise}
\end{cases}
\nonumber \\
\text{where} \nonumber \\
\finalRewF &=\begin{cases}
+10 & ,\ \text{if goal position was reached}\\
-10 \  & ,\ \text{if goal position was not reached}\\
\end{cases}
\label{eq:fuzzyConditions}
\end{align}
where the sub-index A indicates that it is an absorbing state and the sub-index F indicates that it is a \textit{fuzzified} reward. This reward term is defined as:
\begin{align}
\fuzzyRewF &=\termF \oneRewF +( 1-\termF ) \zeroRewF\,, \nonumber\\
\zeroRewF &= - \normF{\touchForceF}\,, \nonumber\\
\oneRewF &= - \fy + \fx -c \,.
\label{eq:fuzzyRew2}
\end{align}
The predicted force vector components are defined by $\fx=\normF{\touchForceF} \sin \angleF$ and $\fy=\touchForceF \cos \angleF $ respectively, and $\angleF$ is the angle between the assistive force and the velocity vector. This angle is calculated through:
\begin{equation}
\angleF =\arccos\left(\frac{\touchForceF \cdot \touchVelVF}{\Vert \touchForceF \Vert \Vert \touchVelVF \Vert }\right)\,.
\label{eq:angleF}
\end{equation}

The term \textit{c} is defined by: $c=\Vert \max (\touchForceF) \Vert$ and the $\termF$ term is defined as:
\begin{equation}
\termF =\begin{cases}
0 & ,\ \text{if}\ \normF{\touchVelVF} = 0\\
1\  & ,\ \text{if}\ \normF{\touchVelVF}  >\extraTermF\\
\frac{\normF{\touchVelVF}}{\extraTermF} & ,\ \text{otherwise}
\end{cases}
\end{equation}

where the scalar positive $x$ term is defined as the maximum user velocity when random forces are applied and $\termF$ term conditions the behavior of the reward function: when the user is performing the task and his current velocity is greater than $\extraTermF$, then the term $\zeroRewF$ is eliminated from the reward function. In this case, the rewards are maximum when assistance forces go in the same direction as user velocity encouraging high assistive forces. 

In contrast, when the user does not perform movements and the velocity is zero, then the $\oneRewF$ term is eliminated from the reward function. Thereby, the assistance forces are not required and reward values are maximum when the forces go to zero. In this context, when the user decreases the velocity less than $\extraTermF$, then, the reward function encourages the DDPG agent to decrease the assistance. In this way, the force magnitude is decreased in a smooth way to provide the user more control on the execution in order to reach the desired position. In addition, using the constant $c$ in $\oneRewF$, ensures that the reward function is always negative, thereby implicitly minimizing the number of required steps.

\begin{algorithm}
	\caption{TAHSC}
	\small{\begin{algorithmic}
\Function{DDPG}{$\discountF, \tau, M, N$}
\State $\criticPVF \leftarrow 0,\ \actorPVF \leftarrow 0,\ \targetCriticPVF \leftarrow 0,\ \targetActorPVF \leftarrow 0,\ \mathcal{R} \leftarrow \emptyset$
\For{$\mathit{episode}=1,M$}
    \State $t \leftarrow 0, \mathcal{T} \leftarrow \emptyset$
    \While{$\sF\ \mathrm{is}\ \mathbf{not}\ \sF_\mathrm{A}$}
        \If{$t=0$}
	        \State $\mathit{features} \leftarrow$\Call{VGG16}{$\mathit{task\ image}$}
        \EndIf
        \State Select action $\actionF$ from $\detParamPolFtime + \ARP$
        \State Execute action $\actionF$, observe reward $\rF_t$ and observe
        \State \hfill new state $\sF_{t+1}$
        \State Store transition $(\sF_{t},\aF_{t},\rF_{t},\sF_{t+1}, 0, \mathit{features})$ in $\mathcal{T}$
        \State $t \leftarrow t+1$
    \EndWhile
    \If{$\mathit{episode}$ was successful}
        \State $\rF_{t-1} \leftarrow \rF_\mathrm{t-1}+10$
        \State Update transition $(\sF_{t-1},\aF_{t-1},\rF_{t-1},\sF_{t}, 1, \mathit{features})$
        \State \hfill in $\mathcal{T}$
    \ElsIf{$\mathit{episode}$ was unsuccessful}
        \State $\rF_{t-1} \leftarrow \rF_{t-1}-10$
        \State Update transition $(\sF_{t-1},\aF_{t-1},\rF_{t-1},\sF_{t}, 1, \mathit{features})$ \State \hfill in $\mathcal{T}$
    \Else
        \State $\mathcal{T} \leftarrow \emptyset$
        \State $\mathit{episode}$ $\leftarrow$ $\mathit{episode}$ $- 1$
    \EndIf
    \State $\bufferF \leftarrow \bufferF \cup \mathcal{T}$
    \For{$t=1,|\mathcal{T}|$}
        \State \Call{Train-minibatch}{$\bufferF, \discountF, \tau, \actorPVF, \targetActorPVF, \criticPVF, \targetCriticPVF, N$}
    \EndFor
\EndFor
\EndFunction


\Statex
\Function{Train-minibatch}{$\bufferF, \discountF, \tau, \actorPVF, \targetActorPVF, \criticPVF, \targetCriticPVF, N$}
    \State Sample a random minibatch $\batchF \subset \bufferF$ of size $N$
    \For{each $b_i: (\sF_i,\aF_i,\rF_i,\nSF_i, \mathrm{A}_i) \in \batchF$}
        \If{$\mathrm{A}_i = 0$}
            \State $y_{i} = \rF_{i} + \gamma \indetParamQF$
        \Else
            \State $y_{i} = \rF_{i} $
        \EndIf
        
    \EndFor
    \State Train $\detParamAppQF$ on all samples $N$
    \State Move $\detParamPolF$ according to the sampled deterministic
    \State \hfill policy gradient
    \State \hfill $\frac{1}{N} \sumaF{1}{N} [\grad_{\aF} \detParamQF \vert_{\sF=\sF_i,\aF=\idetParamPolF} \grad_{\actorPVF} \detParamPolF \vert_{\sF=\sF_i}]$
    \State $\targetActorPVF \leftarrow \targetActorPVF + \tau (\actorPVF - \targetActorPVF)$
    \State $\targetCriticPVF \leftarrow \targetCriticPVF + \tau (\criticPVF - \targetCriticPVF)$
\EndFunction
\end{algorithmic}
}
	\label{alg:ddpg 2_2}
\end{algorithm}

\section{Simulator}\label{Section:simulator}
A simulator of both, the master and slave sides was used to test different network architectures of the proposed HSC-controller, as well as perform a hyper-parameter tuning, and validate the reward function. So, we have an approximation that can be used to find the architecture in which the system achieves better performance.

\subsection{Simulator functioning}
Similar to the real system, the simulator is composed of a master side and a slave side. The master side is responsible for providing the virtual trajectory simulating the user behavior in terms of user position in every time step. To do this, the master-side simulator uses a pre-recorded dataset of \textit{N} successful trajectories which are then used to simulate the current position. 
In contrast, the slave-side simulator is used to provide the virtual manipulator position using the velocity commands coming from the teleoperation controller.

\subsubsection{Master-side simulator}
\begin{figure}
	\centering
	\resizebox{\linewidth}{!}{\tikzset{every picture/.style={line width=0.75pt}} 

\begin{tikzpicture}[x=0.75pt,y=0.75pt,yscale=-1,xscale=1]

\draw [color={rgb, 255:red, 0; green, 0; blue, 0 }  ,draw opacity=1 ][line width=3]  [dash pattern={on 3.38pt off 3.27pt}]  (165,5130) .. controls (292,4834) and (556,4837) .. (686,5133) ;
\draw  [color={rgb, 255:red, 0; green, 0; blue, 0 }  ,draw opacity=1 ][fill={rgb, 255:red, 0; green, 0; blue, 0 }  ,fill opacity=1 ] (162,5130) .. controls (162,5128.29) and (163.34,5126.9) .. (165,5126.9) .. controls (166.66,5126.9) and (168,5128.29) .. (168,5130) .. controls (168,5131.71) and (166.66,5133.1) .. (165,5133.1) .. controls (163.34,5133.1) and (162,5131.71) .. (162,5130) -- cycle ;
\draw  [color={rgb, 255:red, 0; green, 0; blue, 0 }  ,draw opacity=1 ][fill={rgb, 255:red, 0; green, 0; blue, 0 }  ,fill opacity=1 ] (680,5128.2) .. controls (680,5126.49) and (681.34,5125.1) .. (683,5125.1) .. controls (684.66,5125.1) and (686,5126.49) .. (686,5128.2) .. controls (686,5129.91) and (684.66,5131.3) .. (683,5131.3) .. controls (681.34,5131.3) and (680,5129.91) .. (680,5128.2) -- cycle ;
\draw  [color={rgb, 255:red, 0; green, 0; blue, 0 }  ,draw opacity=1 ][fill={rgb, 255:red, 0; green, 0; blue, 0 }  ,fill opacity=1 ] (263,4978.1) .. controls (263,4976.39) and (264.34,4975) .. (266,4975) .. controls (267.66,4975) and (269,4976.39) .. (269,4978.1) .. controls (269,4979.81) and (267.66,4981.2) .. (266,4981.2) .. controls (264.34,4981.2) and (263,4979.81) .. (263,4978.1) -- cycle ;
\draw  [color={rgb, 255:red, 0; green, 0; blue, 0 }  ,draw opacity=1 ][fill={rgb, 255:red, 0; green, 0; blue, 0 }  ,fill opacity=1 ] (240,4911.9) .. controls (240,4910.19) and (241.34,4908.8) .. (243,4908.8) .. controls (244.66,4908.8) and (246,4910.19) .. (246,4911.9) .. controls (246,4913.61) and (244.66,4915) .. (243,4915) .. controls (241.34,4915) and (240,4913.61) .. (240,4911.9) -- cycle ;
\draw  [color={rgb, 255:red, 0; green, 0; blue, 0 }  ,draw opacity=1 ][fill={rgb, 255:red, 0; green, 0; blue, 0 }  ,fill opacity=1 ] (411,4909) .. controls (411,4907.29) and (412.34,4905.9) .. (414,4905.9) .. controls (415.66,4905.9) and (417,4907.29) .. (417,4909) .. controls (417,4910.71) and (415.66,4912.1) .. (414,4912.1) .. controls (412.34,4912.1) and (411,4910.71) .. (411,4909) -- cycle ;
\draw    (243,4915) -- (408,4909.11) ;
\draw [shift={(411,4909)}, rotate = 537.95] [fill={rgb, 255:red, 0; green, 0; blue, 0 }  ][line width=0.08]  [draw opacity=0] (8.93,-4.29) -- (0,0) -- (8.93,4.29) -- cycle    ;
\draw  [color={rgb, 255:red, 0; green, 0; blue, 0 }  ,draw opacity=1 ][fill={rgb, 255:red, 0; green, 0; blue, 0 }  ,fill opacity=1 ] (410,4873.1) .. controls (410,4871.39) and (411.34,4870) .. (413,4870) .. controls (414.66,4870) and (416,4871.39) .. (416,4873.1) .. controls (416,4874.81) and (414.66,4876.2) .. (413,4876.2) .. controls (411.34,4876.2) and (410,4874.81) .. (410,4873.1) -- cycle ;
\draw    (246,4911.9) -- (410.08,4873.78) ;
\draw [shift={(413,4873.1)}, rotate = 526.9200000000001] [fill={rgb, 255:red, 0; green, 0; blue, 0 }  ][line width=0.08]  [draw opacity=0] (8.93,-4.29) -- (0,0) -- (8.93,4.29) -- cycle    ;
\draw  [line width=1.5]  (238.5,4913) .. controls (234.11,4914.59) and (232.72,4917.58) .. (234.31,4921.97) -- (240.5,4938.99) .. controls (242.78,4945.26) and (241.73,4949.19) .. (237.34,4950.78) .. controls (241.73,4949.19) and (245.06,4951.52) .. (247.34,4957.79)(246.31,4954.97) -- (253.53,4974.81) .. controls (255.12,4979.2) and (258.11,4980.59) .. (262.5,4979) ;

\draw (428.5,4928) node  [font=\LARGE] [align=left] {$\displaystyle I^{i}_{t+1}$};
\draw (278.5,4989) node  [font=\LARGE] [align=left] {$\displaystyle I^{i}_{t}$};
\draw (216,4908.5) node  [font=\LARGE]  {$\check{\boldsymbol{p}}_{\mathrm{H}}$};
\draw (140,4954.5) node  [font=\LARGE]  {$\mathnormal{argmin}\left(\mathcal{D}^{i} ,\check{\boldsymbol{p}}_{\mathrm{H}}\right)$};
\draw (683,5144) node  [font=\LARGE] [align=left] {$\displaystyle (\overline{x}_{j} ,\ \overline{y}_{j} ,\ \overline{z}_{j})$};
\draw (162.5,5146) node  [font=\LARGE] [align=left] {$\displaystyle (\overline{x}_{0} ,\ \overline{y}_{0} ,\ \overline{z}_{0})$};
\draw (461.5,4866) node  [font=\LARGE]  {$\check{\boldsymbol{p}}_\mathrm{H}( t+1) \ $};
\draw (332.47,4872.41) node  [font=\LARGE,rotate=-346.81]  {$\check{\boldsymbol{f}}_{\mathrm{U}} +\boldsymbol{f}_{\mathrm{H}}$};
\draw (290.38,4933.35) node  [font=\LARGE,rotate=-358.04]  {$\check{\boldsymbol{f}}_{\mathrm{U}}$};

\end{tikzpicture}}
	\caption{Master-side simulator: Once the reference point w.r.t the sampled trajectory is calculated with the \textit{argmin} function, the virtual user force is obtained through a proportional difference between the desired and the current position.}
	\label{fig:simulator}
\end{figure}
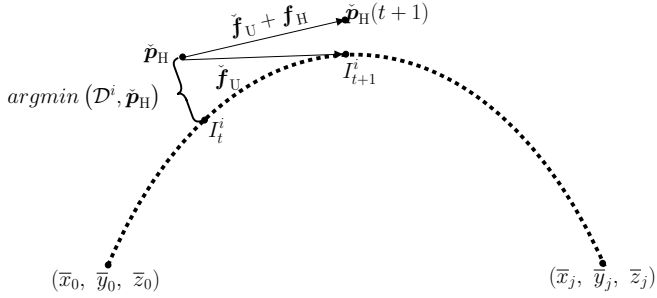

Consider a set of \textit{N} successfully executed demonstrated trajectories:
\begin{equation}
\datasetF=\{\{(x_0^i, y_0^i, z_0^i), ...., (x_{j_i}^i, y_{j_i}^i, z_{j_i}^i)\}^N_{i=1}\},\ {j_i} > 0    \,.
\end{equation}
A single trajectory $\datasetF^i$, $i \in [1,N]$ can be randomly chosen to simulate a single episode for task execution in the following way: the mean of the initial points from the touch $(\Bar{x_0}, \Bar{y_0}, \Bar{z_0})$ of the sampled trajectories is taken as the starting point, and the mean of the final points from the touch $(\Bar{x}_{j_i}, \Bar{y}_{j_i}, \Bar{z}_{j_i})$ is taken as the \textit{goal} point. Then, at time $t$, the current index $I^{i}_{t}$ of the $i$-th trajectory is calculated as the nearest point between the current simulated position $\simTouchPosF(t)$ (hereafter abbreviated to $\simTouchPosF$) and the sampled trajectory $\datasetF^i$.
\begin{equation}
I^{i}_{t} = \arg \min (\datasetF^i, \simTouchPosF)\,.
\label{eq:argmin}
\end{equation}
Then, the index of the next desired point is calculated as the current index added with a \textit{step} constant. 
\begin{equation}
I^{i}_{t+1} = I^{i}_{t} + \mathit{step}\,.
\label{eq:nxIndex}
\end{equation}
So that, given the current simulated position and the next desired sampled position, the simulated user force vector ($\simUserForceF$) is approximated as the difference between the two points multiplied with a proportional gain $k_{u}$.
\begin{align}
\simUserForceF = k_{u} (\datasetF_{I^{i}_{t+1}}^{i} - \simTouchPosF)\,.
\label{eq:userForce}
\end{align}
Moreover, because we are modelling the human behavior, where the hand impedance model has a viscous damper component~\cite{hand_impedance}, we assume that forces are proportional to velocities for small motions deprecating the acceleration~\cite{LAGO2019107}. So, the velocity vector is approximated as the addition between the user force vector $\simUserForceF$ with the assistance forces $\touchForceF$ received from the HSC controller, multiplied with a scaling gain $k_{s}$.
\begin{equation}
\simVelVectorF_{\mathrm{H}} = k_{s} (\simUserForceF + \touchForceF)
\label{eq:touchVel}
\end{equation}
This equation is used to simulate the effect of the application of the combined forces in the system. Finally, this velocity vector is multiplied with the time step $\Delta t$ and added to the current simulated point to get the next simulated point (Euler integration). This process is repeated every time step simulating the virtual trajectory.
\begin{equation}
\simTouchPosF(t+1) = \simTouchPosF + \simVelVectorF_{\mathrm{H}} \Delta t\,.
\label{eq:nxPosition}
\end{equation}
Figure~\ref{fig:simulator} illustrates the basic functioning of the master-side simulator. In this way, the Master-side simulator provides a position vector $\simTouchPosF$ and receives the assistance forces $\touchForceF$ from the controller.

\subsubsection{Slave-side simulator}
On the other hand, the functioning of the Slave-side simulator is more simple. Since the slave side \textit{replicates} the movements of the master side, only the velocity commands and the initial and final points from the manipulator are necessary. Similarly to the Master-side simulator, the mean of the samples' initial points $(\Bar{x_0}, \Bar{y_0}, \Bar{z_0})$ and the mean of the samples' final points $(\Bar{x_j}, \Bar{y_j}, \Bar{z_j})$  can be calculated from a set of demonstrated trajectories for the manipulator case becoming in the initial position and the \textit{goal} position. 

Then, at time $t$, given a certain position for the manipulator  $\simDobotPosF$, the next position can be calculated simply by using the Euler integration method, similar to~(\ref{eq:nxPosition}). Thereby, the virtual manipulator position $\simDobotPosF$ is sent to the HSC controller. Similar with the master-side simulator, this process is repeated every time step until the current manipulator position reaches the \textit{goal} point, where the episode ends and the trajectory is restarted.

\begin{equation}
\simDobotPosF(t+1) = \simDobotPosF + \simVelVectorF_{\mathrm{M}} \Delta t\,.
\label{eq:nxPositionDobot}
\end{equation}

The architecture for the simulator is similar as Figure~\ref{fig:teleoperation camera input}. Note that since we do not simulate the camera, the task is therefore taken from the demonstration trajectory and the direction is directly passed to the DDPG network.


\section{Experimental Setup}

\begin{figure}
	\centering
	\subfigure[][\label{fig:touch}]{\includegraphics[width=.4\linewidth]{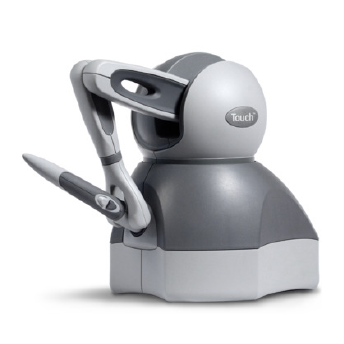}}
	\subfigure[][\label{fig:dobot}]{\includegraphics[width=.49\linewidth]{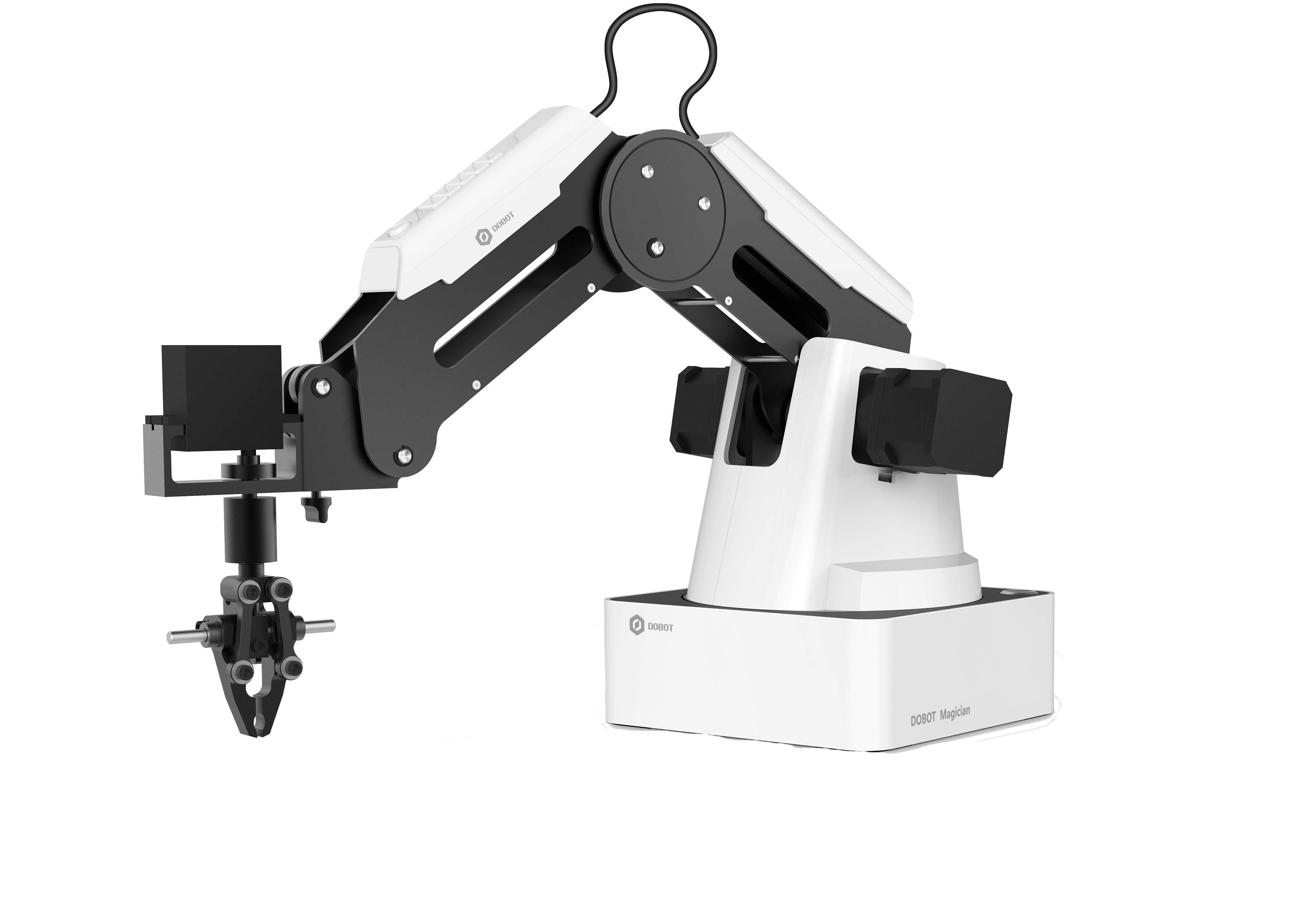}}
	\caption{Input/output devices. \subref{fig:touch}: Touch Haptic Device used in the master side.
	\subref{fig:dobot}: Dobot Magician robotic arm used in the slave-side.
	}
	\label{fig:dobot_magician} 
\end{figure}

The implemented system is composed of three components: the master side, the central controller and the slave side. For the real system, the master side is composed of the 6 DOF Touch haptic device from 3D Systems, Inc. (Figure~\ref{fig:touch}), while on the slave side the 4 DOF robotic manipulator Dobot Magician from DOBOT (Figure~\ref{fig:dobot}). On the other hand, for the simulation, the master and the slave side were replaced by two programs implemented following the specifications given in Section~\ref{Section:simulator}. For both implementations, the real system and the simulator, a computer was used as a central controller. This computer is a desktop workstation with a 4-core Intel® Core™ i7-7700 CPU @ 3.60GHz as the processor, 16 GB of RAM, GeForce GTX 1080 as graphic card and Ubuntu 18.04.3 LTS as operating system. The Robot Operating System (ROS) Melodic was the middleware used for communication between the various software components in the system. 
The ROS controller for the Touch device and the Dobot Magician were adapted from \cite{touchROSpackage} and \cite{dobotROSpackage} respectively. The ANN used in this research were implemented using the Keras libraries and trained using the Tensorflow back-end. Finally, to capture the visual information a C922 Pro Stream Webcam from Logitech was used.

\section{Simulation analysis and preliminary tests}\label{section:preliminary_tests}

As preliminary tests, the simulator was first applied in order to find the hyperparameters to be used in the real system implementation. Following the specifications given in previous sections, different experiments were performed with the simulator varying the different hyperparameters for the DDPG network: batch size, discount rate $\gamma$, and the update target parameter $\tau$. In the same way, to perform the action exploration, \arp\ process~\cite{Korenkevych:2019:APC:3367243.3367422} was used and its parameters were also varied: the order $p$ and the parameter $\alpha$. The complete set of parameters and the variations are summarized in Table~\ref{table:master simulator parameters} and~\ref{table:DDPG0 simulator parameters}. On the other hand,  in the simulations, the task detection was performed selecting the task automatically at the beginning of the episodes. 
In addition, as every episode of the simulation is composed of the task execution in one direction, and, since the task is being performed in both directions, we consider two episodes as an epoch. An extra consideration was the addition of a testing epoch every 10 training epochs and the condition that the simulator always completes the task successfully. This condition remains in the fact that the master side simulator bases its trajectory in a set of successful demonstrations. For this reason, in the final part of the execution of the simulated trajectory, its final position is always proximal to the goal position, which leads to a successful execution.

\begin{table}
	\centering
	\caption{Parameters used for master-side simulator. These values were obtained using a similar technique as in Table~\ref{table:DDPG0 simulator parameters}}
	\begin{tabular}{cc}
    \hline 
    \textbf{Parameter} & \textbf{Value} \\
    \hline 
    Step                    & 12\\
    User gain ($k_u$)       & 5\\
    Force gain ($k_s$)      & 5 \\
    Sampling time ($\Delta t$) & 0.01 \\
    \hline
\end{tabular}
	\label{table:master simulator parameters}
\end{table}
\begin{table}
	\centering
	\caption{Parameters used for simulation.}
	\begin{tabular}{cccc}
    \hline 
    \textbf{Parameter} & \textbf{Min} & \textbf{Max} & \textbf{Best value} \\
    \hline 
    Batch size ($N$)                &   32  & 128      & 64\\
    Discount rate ($\gamma$)        &   0.96 & 0.99     & 0.99\\
    Update target parameter ($\tau$)&   0.001 & 0.01    & 0.01 \\
    Actor optimizer                 & - & - & Adam \\
    Actor learning rate             & 0.0001 & 0.001 & 0.001 \\
    Critic optimizer                & - & - & Adam \\
    Critic learning rate            & 0.0001 & 0.001 & 0.0001 \\
    Epochs                          & 15 & 100 & 25 \\
    \arp\ order ($p$)               & 2 & 3         & 3 \\
    \arp\ parameter ($\alpha$)      & 0.8 & 0.9     & 0.8 \\
    \hline
\end{tabular}
	\label{table:DDPG0 simulator parameters}
\end{table}
\subsection{Results}
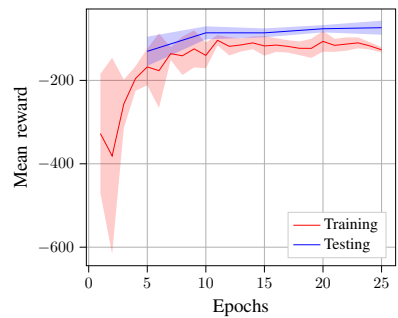
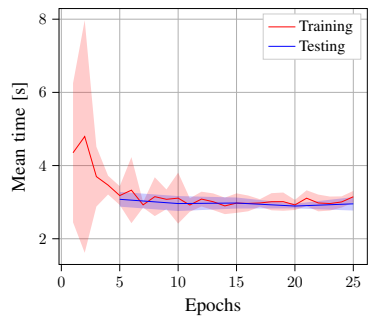
\begin{figure}
	\centering
	\subfigure[][Epoch rewards.\label{fig:simDDPGbaseReward}]{\resizebox{0.59\linewidth}{!}{
\begin{tikzpicture}

\begin{axis}[
legend cell align={left},
legend style={at={(0.97,0.03)}, anchor=south east, draw=white!80.0!black},
tick align=outside,
tick pos=left,
x grid style={lightgray!92.0261437908!black},
xlabel={\large Epochs},
xmajorgrids,
xmin=-0.2, xmax=26.2,
xtick style={color=black},
y grid style={lightgray!92.0261437908!black},
ylabel={\large Mean reward},
ymajorgrids,
ymin=-644.443777020821, ymax=-28.7488845714924,
ytick style={color=black}
]
\path [fill=red, fill opacity=0.2]
(axis cs:1,-470.707866938314)
--(axis cs:1,-184.412299935543)
--(axis cs:2,-145.744597861834)
--(axis cs:3,-198.291616642731)
--(axis cs:4,-165.8744518547)
--(axis cs:5,-123.524020040924)
--(axis cs:6,-87.3001568659033)
--(axis cs:7,-120.376322659211)
--(axis cs:8,-94.8889969658153)
--(axis cs:9,-79.9112678234833)
--(axis cs:10,-108.997171761842)
--(axis cs:11,-91.2541144150404)
--(axis cs:12,-96.4291520403908)
--(axis cs:13,-99.1253336300865)
--(axis cs:14,-94.9306650040001)
--(axis cs:15,-93.9958019338447)
--(axis cs:16,-98.8404884453744)
--(axis cs:17,-103.509373310707)
--(axis cs:18,-106.547443573091)
--(axis cs:19,-100.222381601596)
--(axis cs:20,-81.6375663786651)
--(axis cs:21,-100.436299567349)
--(axis cs:22,-96.8906739107633)
--(axis cs:23,-97.0870886222962)
--(axis cs:24,-108.351104637183)
--(axis cs:25,-120.358563612108)
--(axis cs:25,-132.910878988425)
--(axis cs:25,-132.910878988425)
--(axis cs:24,-126.133606525905)
--(axis cs:23,-122.987208516095)
--(axis cs:22,-128.915141839302)
--(axis cs:21,-131.581574990819)
--(axis cs:20,-131.258226038138)
--(axis cs:19,-146.093670238525)
--(axis cs:18,-139.539020487908)
--(axis cs:17,-133.386136906748)
--(axis cs:16,-131.703070075408)
--(axis cs:15,-140.77872388287)
--(axis cs:14,-125.147321870765)
--(axis cs:13,-130.8089762957)
--(axis cs:12,-140.311528461341)
--(axis cs:11,-116.433010043849)
--(axis cs:10,-170.96375727503)
--(axis cs:9,-169.014014482835)
--(axis cs:8,-187.06623991442)
--(axis cs:7,-151.404034054182)
--(axis cs:6,-265.940656408166)
--(axis cs:5,-211.875207199436)
--(axis cs:4,-224.842405216914)
--(axis cs:3,-314.318986624769)
--(axis cs:2,-616.457645545852)
--(axis cs:1,-470.707866938314)
--cycle;

\path [fill=blue, fill opacity=0.2]
(axis cs:5,-165.440895552233)
--(axis cs:5,-95.1847647021762)
--(axis cs:10,-70.3459432928981)
--(axis cs:15,-75.3175259406007)
--(axis cs:20,-67.355762781822)
--(axis cs:25,-56.7350160464619)
--(axis cs:25,-90.4781882322093)
--(axis cs:25,-90.4781882322093)
--(axis cs:20,-84.9650555415672)
--(axis cs:15,-96.6609543917832)
--(axis cs:10,-101.488232830143)
--(axis cs:5,-165.440895552233)
--cycle;

\addplot [semithick, red]
table {%
1 -327.560083436928
2 -381.101121703843
3 -256.30530163375
4 -195.358428535807
5 -167.69961362018
6 -176.620406637035
7 -135.890178356697
8 -140.977618440118
9 -124.462641153159
10 -139.980464518436
11 -103.843562229445
12 -118.370340250866
13 -114.967154962893
14 -110.038993437382
15 -117.387262908358
16 -115.271779260391
17 -118.447755108728
18 -123.043232030499
19 -123.15802592006
20 -106.447896208402
21 -116.008937279084
22 -112.902907875033
23 -110.037148569196
24 -117.242355581544
25 -126.634721300267
};
\addlegendentry{Training}
\addplot [semithick, blue]
table {%
5 -130.312830127205
10 -85.9170880615207
15 -85.9892401661919
20 -76.1604091616946
25 -73.6066021393356
};
\addlegendentry{Testing}
\end{axis}

\end{tikzpicture}}}\\
	\subfigure[][Time per epoch.\label{fig:simDDPGbaseStep}]{\resizebox{0.55\linewidth}{!}{
\begin{tikzpicture}

\begin{axis}[
legend cell align={left},
legend style={draw=white!80.0!black},
tick align=outside,
tick pos=left,
x grid style={lightgray!92.0261437908!black},
xlabel={\large Epochs},
xmajorgrids,
xmin=-0.2, xmax=26.2,
xtick style={color=black},
y grid style={lightgray!92.0261437908!black},
ylabel={\large Mean time [s]},
ymajorgrids,
ymin=1.29703921220877, ymax=8.29096078779123,
ytick style={color=black}
]
\path [fill=red, fill opacity=0.2]
(axis cs:1,2.44974721096881)
--(axis cs:1,6.25025278903119)
--(axis cs:2,7.97305526162839)
--(axis cs:3,4.51628063900832)
--(axis cs:4,3.72303720442576)
--(axis cs:5,3.43643938920408)
--(axis cs:6,4.23112086076983)
--(axis cs:7,3.00096535683399)
--(axis cs:8,3.68075287564862)
--(axis cs:9,3.34689004710393)
--(axis cs:10,3.80650655308354)
--(axis cs:11,3.10556256624539)
--(axis cs:12,3.28628860310153)
--(axis cs:13,3.23950576645886)
--(axis cs:14,3.12342042670573)
--(axis cs:15,3.2407298986084)
--(axis cs:16,3.18129417704817)
--(axis cs:17,3.08073676716985)
--(axis cs:18,3.24640686451764)
--(axis cs:19,3.26334321998235)
--(axis cs:20,3.06255346996899)
--(axis cs:21,3.32460629184849)
--(axis cs:22,3.20731488317026)
--(axis cs:23,3.15374164118175)
--(axis cs:24,3.15873923332677)
--(axis cs:25,3.30747036405863)
--(axis cs:25,2.98852963594137)
--(axis cs:25,2.98852963594137)
--(axis cs:24,2.85326076667323)
--(axis cs:23,2.77025835881825)
--(axis cs:22,2.74868511682974)
--(axis cs:21,2.89139370815151)
--(axis cs:20,2.78544653003101)
--(axis cs:19,2.76465678001765)
--(axis cs:18,2.77759313548236)
--(axis cs:17,2.87526323283015)
--(axis cs:16,2.75070582295183)
--(axis cs:15,2.7072701013916)
--(axis cs:14,2.67257957329427)
--(axis cs:13,2.78449423354114)
--(axis cs:12,2.88171139689847)
--(axis cs:11,2.73843743375461)
--(axis cs:10,2.41349344691646)
--(axis cs:9,2.80910995289607)
--(axis cs:8,2.61924712435138)
--(axis cs:7,2.85103464316601)
--(axis cs:6,2.42487913923017)
--(axis cs:5,2.92356061079592)
--(axis cs:4,3.20896279557424)
--(axis cs:3,2.87971936099168)
--(axis cs:2,1.61494473837161)
--(axis cs:1,2.44974721096881)
--cycle;

\path [fill=blue, fill opacity=0.2]
(axis cs:5,2.87694077036368)
--(axis cs:5,3.27905922963632)
--(axis cs:10,3.16655212371352)
--(axis cs:15,3.12679442222724)
--(axis cs:20,2.95883091262267)
--(axis cs:25,3.13730511511414)
--(axis cs:25,2.77069488488586)
--(axis cs:25,2.77069488488586)
--(axis cs:20,2.82916908737733)
--(axis cs:15,2.82520557777276)
--(axis cs:10,2.75744787628648)
--(axis cs:5,2.87694077036368)
--cycle;

\addplot [semithick, red]
table {%
1 4.35
2 4.794
3 3.698
4 3.466
5 3.18
6 3.328
7 2.926
8 3.15
9 3.078
10 3.11
11 2.922
12 3.084
13 3.012
14 2.898
15 2.974
16 2.966
17 2.978
18 3.012
19 3.014
20 2.924
21 3.108
22 2.978
23 2.962
24 3.006
25 3.148
};
\addlegendentry{Training}
\addplot [semithick, blue]
table {%
5 3.078
10 2.962
15 2.976
20 2.894
25 2.954
};
\addlegendentry{Testing}
\end{axis}

\end{tikzpicture}}}
	\caption{Learning curves obtained for simulation tests.
	}
	\label{fig:DDPG simulation graphs}
\end{figure}

Figure \ref{fig:DDPG simulation graphs} shows the results obtained for preliminary tests. These graphs are a mean of a set of five simulations. The solid line represents the mean value and the shaded area represents the confidence interval for rewards and time graphs respectively, obtaining in the last testing epoch a mean reward of $-73.607$ with a mean of $2.954$ seconds per epoch.
\subsection{Discussion}
As the results of the preliminary tests simulated for the proposed system show, the implemented algorithm achieves convergence in a limited amount of episodes. Practically, after 15 epochs approximately the DDPG agent is able to learn an acceptable policy presenting stable functioning in training and testing episodes (Figure~\ref{fig:simDDPGbaseReward}). This was confirmed observing the necessary amount of time required to perform the task shown in Figure~\ref{fig:simDDPGbaseStep}, where it can be observed that the time decreases along the training with respect to the first episodes.

In addition, the learning curves in the presented graphs illustrate stable functioning for the agent in the final episodes of the training. Finally, it can be concluded that the use of the proposed system presents a stable convergence in the learning of the virtual guiding forces to assist users in the \textit{pick-and-place} task realization.


\section{EXPERIMENTAL ANALYSIS}\label{sect:results}

The previous section showed preliminary results demonstrating that the learned policy of the DDPG algorithm in the teleoperated simulated system presents stable convergence. Those results serve as a basis for the application of the proposed method in a real system. This section presents the implementation and results in the real system described in Section~\ref{section:method}, validating the obtained results. Finally, a set of experiments using different subjects was carried out in order to validate the implemented controller.

\subsection{Preliminary considerations}

For the real system setup, some considerations were added in contrast with the simulator. First of all, the main difference lies in the fact that the simulations always finish in a successful episode, however, in the real system, this condition does not always hold. For that reason, it was necessary to implement a conditional function that described the episode behavior as similar as~(\ref{eq:fuzzyConditions}). Three conditions were considered: successful task completion, unsuccessful task completion and finally, task fail and repeat episode. \textit{Successful task completion} refers to when the user completes the task placing the object in the goal position. \textit{Unsuccessful task completion} refers to when the user completes the task but places the object in a wrong position. Finally, if the ball is dropped or an unexpected event occurs, then it is considered that the task fails and the episode needs to be restarted.


Thereby, after performing every episode, the operator was asked to select the option that best described his performance in that execution. In this context, these conditions were added in the $\finalRewF$ term of the absorbing state in~(\ref{eq:fuzzyConditions}), assigning different rewards according to the described conditions; that is, if the user chooses the first option, an extra high reward $(+10)$ is assigned for the absorbing state. In contrast, if it was an unsuccessful episode and the user chooses the second option, a high penalty reward $(-10)$ is assigned to the absorbing state in that episode. Finally, if the third option is selected, then the episode is repeated from the last start position and the collected samples in the current episode are discarded from the replay buffer memory.

On the other hand, to prevent possible damage to the device, the output forces were clipped in a safe range as following:
\begin{equation}
\touchForceFmin \leq \touchForceF \leq \touchForceFmax\,.
\label{eq:variable forces}
\end{equation}
where the maximum and minimum values were set according to the Haptic device manufacturer~\cite{userGuideTouch}.
As a final consideration, similar to the simulator, an epoch was considered as a set of two episodes, a different task direction for each one. 
Table~\ref{table:ss_detection_variable_fuzzy} summarizes the parameters used in this experiment based on the parameters found in the previous section.

\begin{table}
	\centering
	\caption{Parameters used for the real tests.}
	\begin{tabular}{cc}
	\hline 
	\textbf{Parameter} & \textbf{Value} \\
	\hline 
	Batch size   ($N$)                       & 64\\
	Discount rate  ($\gamma$)          & 0.99\\
	Update target parameter ($\tau$)    & 0.01 \\
	Epochs                            & 25 \\
	\arp\ order ($p$)              & 3 \\
	\arp\ parameter ($\alpha$)   & 0.9 \\
	\hline
\end{tabular}
	\label{table:ss_detection_variable_fuzzy}
\end{table}

\subsection{Results}
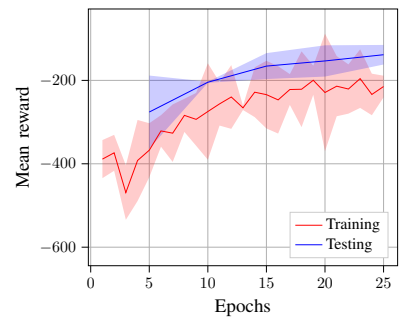
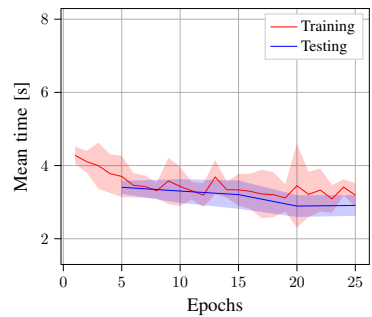
\begin{figure}
	\centering
	\subfigure[][Epoch rewards.\label{fig:SSTDFvrewards}]{\resizebox{0.59\linewidth}{!}{
\begin{tikzpicture}

\begin{axis}[
legend cell align={left},
legend style={at={(0.97,0.03)}, anchor=south east, draw=white!80.0!black},
tick align=outside,
tick pos=left,
x grid style={lightgray!92.0261437908!black},
xlabel={\large Epochs},
xmajorgrids,
xmin=-0.2, xmax=26.2,
xtick style={color=black},
y grid style={lightgray!92.0261437908!black},
ylabel={\large Mean reward},
ymajorgrids,
ymin=-644.443777020821, ymax=-28.7488845714924,
ytick style={color=black}
]
\path [fill=red, fill opacity=0.2]
(axis cs:1,-434.501911190648)
--(axis cs:1,-343.119513495139)
--(axis cs:2,-330.547988806065)
--(axis cs:3,-403.671719902112)
--(axis cs:4,-294.83106164023)
--(axis cs:5,-302.684608301796)
--(axis cs:6,-283.373956207739)
--(axis cs:7,-257.599451302611)
--(axis cs:8,-244.100187117643)
--(axis cs:9,-230.09367305787)
--(axis cs:10,-158.988803062667)
--(axis cs:11,-204.313145811732)
--(axis cs:12,-163.380939129145)
--(axis cs:13,-260.510462588941)
--(axis cs:14,-168.572604220104)
--(axis cs:15,-153.068666700644)
--(axis cs:16,-165.956683918362)
--(axis cs:17,-184.929968663706)
--(axis cs:18,-130.373966723673)
--(axis cs:19,-164.760131158408)
--(axis cs:20,-87.995764295943)
--(axis cs:21,-141.746598707923)
--(axis cs:22,-161.582641910375)
--(axis cs:23,-125.534240555635)
--(axis cs:24,-183.892302065594)
--(axis cs:25,-188.694399030137)
--(axis cs:25,-240.701749032382)
--(axis cs:25,-240.701749032382)
--(axis cs:24,-283.557040545326)
--(axis cs:23,-265.977379876745)
--(axis cs:22,-279.783242674474)
--(axis cs:21,-286.037435674679)
--(axis cs:20,-370.050948642535)
--(axis cs:19,-234.607206802736)
--(axis cs:18,-312.100535506325)
--(axis cs:17,-258.850055644213)
--(axis cs:16,-327.548802946965)
--(axis cs:15,-315.256831453788)
--(axis cs:14,-288.078607952473)
--(axis cs:13,-271.015192450969)
--(axis cs:12,-316.206413256244)
--(axis cs:11,-308.377879106674)
--(axis cs:10,-390.3158143727)
--(axis cs:9,-356.817942815606)
--(axis cs:8,-323.66637311527)
--(axis cs:7,-395.569562727667)
--(axis cs:6,-359.000339153115)
--(axis cs:5,-432.258307181372)
--(axis cs:4,-489.393453745672)
--(axis cs:3,-534.326527413024)
--(axis cs:2,-416.774356202328)
--(axis cs:1,-434.501911190648)
--cycle;

\path [fill=blue, fill opacity=0.2]
(axis cs:5,-363.434398852774)
--(axis cs:5,-188.362243095362)
--(axis cs:10,-203.029648177522)
--(axis cs:15,-134.584772333259)
--(axis cs:20,-115.833098831141)
--(axis cs:25,-114.930411861278)
--(axis cs:25,-161.949326045418)
--(axis cs:25,-161.949326045418)
--(axis cs:20,-190.851161941772)
--(axis cs:15,-196.844035475839)
--(axis cs:10,-205.815239829059)
--(axis cs:5,-363.434398852774)
--cycle;

\addplot [semithick, red]
table {%
1 -388.810712342894
2 -373.661172504196
3 -468.999123657568
4 -392.112257692951
5 -367.471457741584
6 -321.187147680427
7 -326.584507015139
8 -283.883280116456
9 -293.455807936738
10 -274.652308717684
11 -256.345512459203
12 -239.793676192695
13 -265.762827519955
14 -228.325606086288
15 -234.162749077216
16 -246.752743432663
17 -221.89001215396
18 -221.237251114999
19 -199.683668980572
20 -229.023356469239
21 -213.892017191301
22 -220.682942292425
23 -195.75581021619
24 -233.72467130546
25 -214.698074031259
};
\addlegendentry{Training}
\addplot [semithick, blue]
table {%
5 -275.898320974068
10 -204.422444003291
15 -165.714403904549
20 -153.342130386457
25 -138.439868953348
};
\addlegendentry{Testing}
\end{axis}

\end{tikzpicture}}}\\
	\subfigure[][Time per epoch.\label{fig:SSTDFvsteps}]{\resizebox{0.55\linewidth}{!}{
\begin{tikzpicture}

\begin{axis}[
legend cell align={left},
legend style={draw=white!80.0!black},
tick align=outside,
tick pos=left,
x grid style={lightgray!92.0261437908!black},
xlabel={\large Epochs},
xmajorgrids,
xmin=-0.2, xmax=26.2,
xtick style={color=black},
y grid style={lightgray!92.0261437908!black},
ylabel={\large Mean time [s]},
ymajorgrids,
ymin=1.29703921220877, ymax=8.29096078779123,
ytick style={color=black}
]
\path [fill=red, fill opacity=0.2]
(axis cs:1,4.04457681108079)
--(axis cs:1,4.51542318891921)
--(axis cs:2,4.4055187493346)
--(axis cs:3,4.62954902664298)
--(axis cs:4,4.30766675269403)
--(axis cs:5,4.2686081868789)
--(axis cs:6,3.79539219039738)
--(axis cs:7,3.72987648231156)
--(axis cs:8,3.53240061465796)
--(axis cs:9,4.21460730828302)
--(axis cs:10,3.96445969971261)
--(axis cs:11,3.55363808274479)
--(axis cs:12,3.51745417074683)
--(axis cs:13,4.15553542281286)
--(axis cs:14,3.5625329675731)
--(axis cs:15,3.77141983442494)
--(axis cs:16,3.77659625571474)
--(axis cs:17,3.88756692845637)
--(axis cs:18,3.81647572894998)
--(axis cs:19,3.48794773216986)
--(axis cs:20,4.59520065538694)
--(axis cs:21,3.82941050517273)
--(axis cs:22,3.92043557679018)
--(axis cs:23,3.46683378630313)
--(axis cs:24,3.62956572857044)
--(axis cs:25,3.51542416640905)
--(axis cs:25,2.87124250025762)
--(axis cs:25,2.87124250025762)
--(axis cs:24,3.19043427142956)
--(axis cs:23,2.71316621369687)
--(axis cs:22,2.73956442320982)
--(axis cs:21,2.59725616149394)
--(axis cs:20,2.30479934461306)
--(axis cs:19,2.74538560116348)
--(axis cs:18,2.59019093771669)
--(axis cs:17,2.5590997382103)
--(axis cs:16,2.81673707761859)
--(axis cs:15,2.9019134989084)
--(axis cs:14,3.1174670324269)
--(axis cs:13,3.21779791052048)
--(axis cs:12,2.86921249591983)
--(axis cs:11,3.06636191725521)
--(axis cs:10,2.89554030028739)
--(axis cs:9,2.94539269171698)
--(axis cs:8,3.09426605200871)
--(axis cs:7,3.12345685102178)
--(axis cs:6,3.12460780960262)
--(axis cs:5,3.13805847978777)
--(axis cs:4,3.25233324730597)
--(axis cs:3,3.36378430669035)
--(axis cs:2,3.80781458399873)
--(axis cs:1,4.04457681108079)
--cycle;

\path [fill=blue, fill opacity=0.2]
(axis cs:5,3.22800281835722)
--(axis cs:5,3.58533051497611)
--(axis cs:10,3.62846441996262)
--(axis cs:15,3.59394420948672)
--(axis cs:20,3.19218541600127)
--(axis cs:25,3.19461275712487)
--(axis cs:25,2.62538724287513)
--(axis cs:25,2.62538724287513)
--(axis cs:20,2.5944812506654)
--(axis cs:15,2.81938912384661)
--(axis cs:10,2.97820224670405)
--(axis cs:5,3.22800281835722)
--cycle;

\addplot [semithick, red]
table {%
1 4.28
2 4.10666666666667
3 3.99666666666667
4 3.78
5 3.70333333333333
6 3.46
7 3.42666666666667
8 3.31333333333333
9 3.58
10 3.43
11 3.31
12 3.19333333333333
13 3.68666666666667
14 3.34
15 3.33666666666667
16 3.29666666666667
17 3.22333333333333
18 3.20333333333333
19 3.11666666666667
20 3.45
21 3.21333333333333
22 3.33
23 3.09
24 3.41
25 3.19333333333333
};
\addlegendentry{Training}
\addplot [semithick, blue]
table {%
5 3.40666666666667
10 3.30333333333333
15 3.20666666666667
20 2.89333333333333
25 2.91
};
\addlegendentry{Testing}
\end{axis}

\end{tikzpicture}}}
	\caption{Learning curves and trajectory obtained for the TAHSC controller.}
	\label{fig:DDPG2_unnormalized_fuzzy}
\end{figure}

Results for the use of this approach are shown in Figure~\ref{fig:DDPG2_unnormalized_fuzzy} showing the learning curve and the amount of the time for a set of three experiments. The solid line represents the mean value and the shaded area represents the confidence interval respectively, obtaining in the last testing epoch a mean reward of $-138.439$ with a mean of $2.91$ seconds per epoch. 

\subsection{Discussion}

\begin{figure}
	\centering 
	\includegraphics[width=0.3\textwidth]{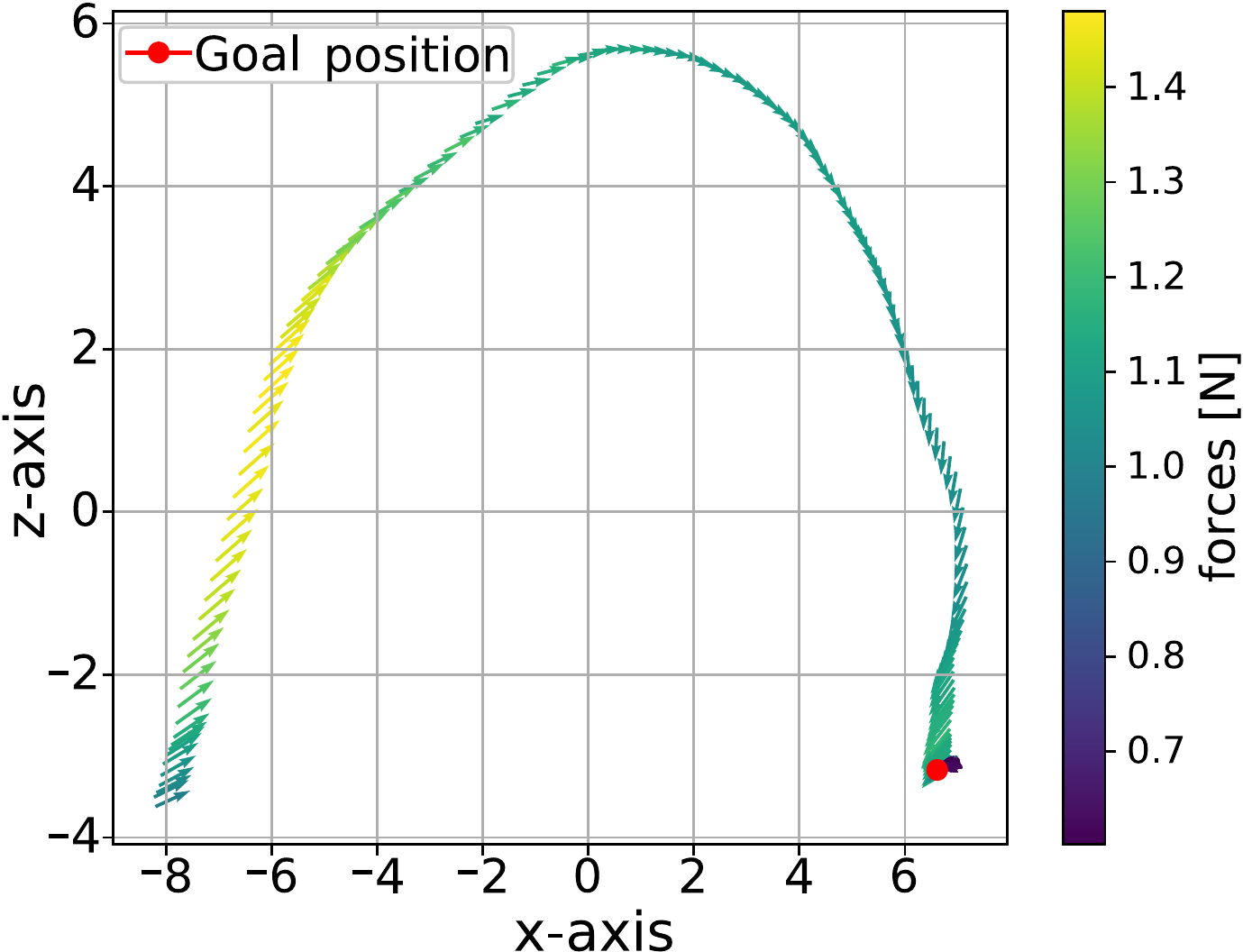}
	\caption{Trajectory obtained with the TAHSC controller.}
	\label{fig:good_behavior} 
\end{figure}

As is observed in the learning curves, the set of experiments performed for the TAHSC controller presents similar results with the simulations. Another observable result is presented in Figure~\ref{fig:good_behavior} where it can be seen that the forces decrease when goal position is being reached by the user. Thereby, the functioning of the reward function encouraging the agent to learn lower assistive forces and gives more control is demonstrated. As expected, similar to previous simulations, the DDPG agent presents a stable convergence in the final part of the training in terms of rewards and time, which leads the next conclusion: the TAHSC controller can be successfully applied in teleoperated systems providing helping to assisting users in the task completion. Finally, it is worth mentioning that all the experiments until now were performed by the same person, for future references named as Subject~0.


\begin{figure*}
	\centering
	\includegraphics[width=0.8\textwidth]{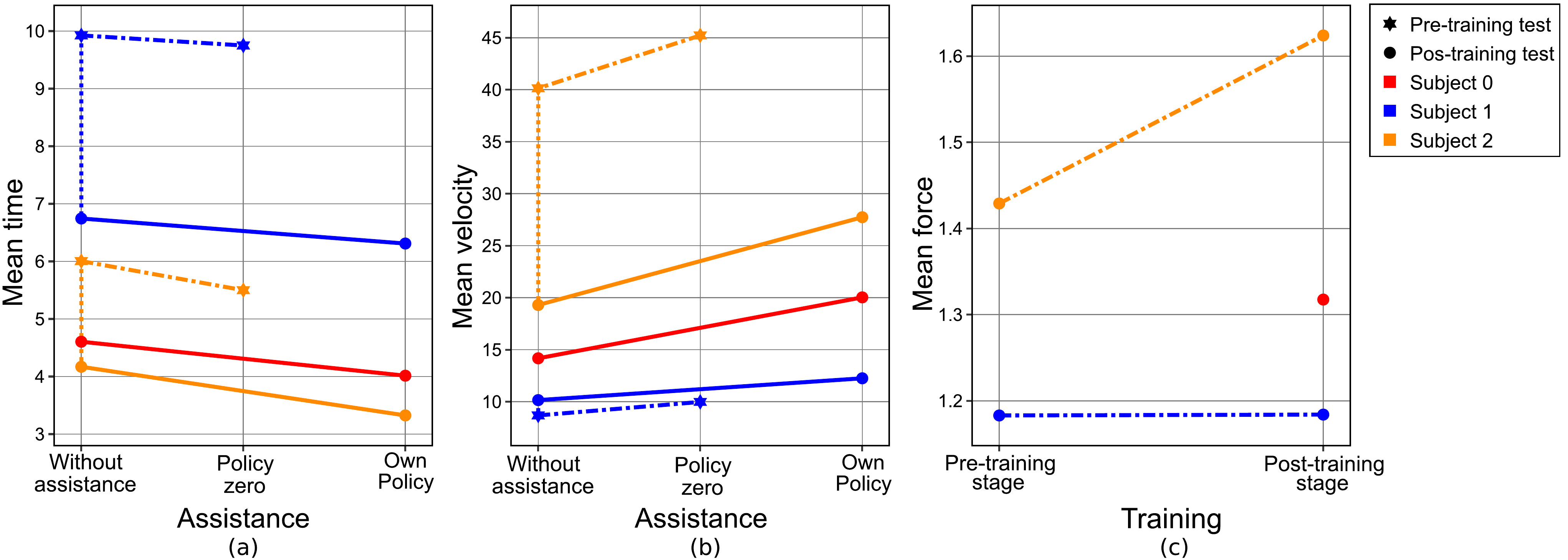}
	\caption{Results for the different subjects. (a): Mean time per epoch for test subjects. The dash lines represent the performance on the pre-training tests, the dot lines represent the performance on the policy training stage and the solid lines represent the performance on the pos-training tests. (b): Velocity results for test subjects. The dash lines represent the performance on the pre-training tests, the dot lines represent the performance on the policy training stage and the solid lines represent the performance on the pos-training tests. (c): Force result for test subjects. The dash lines represent the performance from the pre-training tests to the pos-training tests.
	}
	\label{fig:results_subjects} 
\end{figure*}

\section{Validation with different subjects}\label{Different_subjects_tests}

To investigate the functioning of the developed TAHSC controller, two different subjects were asked to perform a series of tests. These tests consisted of four stages: familiarity with the system, pre-training tests, policy training and post-training tests. In the first stage, the basic functioning of the implemented system was explained while the testing subject was manipulating the devices using the teleoperation controller. Next, a set of epochs were tested with and without providing assistance with a trained policy by Subject~0 (named as \textit{policy-zero}) while the testing subject was trying to perform the task. Then, the testing subject was asked to train a new policy with the RL-controller performing the designed task. Finally, a new set of tests with the same structure as the pre-training stage was performed with the trained policy.

\subsection{Results}
Results showing the mean time per epochs for all subjects on the pre-training and post-training stages and with and without assistance  are presented in Figure~\ref{fig:results_subjects}(a). In addition, results for the mean velocity and force are presented in Figures~\ref{fig:results_subjects}(b) and Figure~\ref{fig:results_subjects}(c) respectively.

\subsection{Discussion}
Various conclusions can be drawn from the results presented in this section. First of all, Figure~\ref{fig:results_subjects} reflects different behaviors for every subject when performing the task in the different situations. From Subfigure~\ref{fig:results_subjects}(a), we can see that all subjects complete the task more quickly when receiving generic assistance before training (dashed lines). However, a larger difference is found when the subjects trained their own policy (solid lines), indicating that a personalized assistance policy significantly reduces the completion time ($p < 0.05$). Another interesting result observed in Subfigure~\ref{fig:results_subjects}(a) is that the mean time required to complete the task without assistance decreases in all cases comparing before and after performing the training (dotted line), which indicates that the subjects are able to learn the task independent of the received assistance.

On the other hand, from Subfigures~\ref{fig:results_subjects}(b) and~\ref{fig:results_subjects}(c), we can say for Subject~1 that despite the little change in the mean velocity and force on the pre-training and post-training stages, the mean time required to perform the task is completely different (Subfigure~\ref{fig:results_subjects}(a)), in contrast with the results obtained for Subject~2 where we can see a large difference in the mean velocity and force before and after performing the training (dashed and solid lines). In this context, an observable result that confirms the unique behavior for each user is when velocity and time are compared. For example, Subject~2 presents the highest velocity in all situations, which results in less time for all cases. On the other hand, opposite performance is achieved by Subject~1, who presented the lowest velocity and required more time. Finally, we can see that Subject~0 presents an intermediate behavior compared with the other subjects.

These results show that, although the assistive policy learned through training with one user can help other subjects, optimal performance can only be achieved by training a user-specific policy. In a common RL-implementation, the observations and rewards received by the RL agent are a result of his direct interaction with the environment through the taken actions. However, in RL-implementations with a human-in-the-loop, the interaction between the RL agent and the environment is through the user. As result, the RL agent optimizes the combined user-environment system. Therefore, if the user behaves differently, a different policy will be optimal. Thereby the policy becomes particular for every user.

\section{Conclusions and future work}
In this work, a novel HSC controller was developed to be used in a teleoperated system composed by a robotic arm and a haptic device. The proposed controller is composed of an RL section and a direct teleoperation section. Taking advantage of the policy gradient methods that present stable functioning on continuous control systems, the RL-controller was implemented using as core the DDPG algorithm. On the other side, a proportional controller was used as the base of the teleoperation controller.

The resulting HSC controller was able to learn custom guiding forces for all subjects who tested the system (Section~\ref{Different_subjects_tests}). Thereby, the learned policy becomes a personal policy which contains enough guidelines to assist the particular behavior of the user it was trained for. Besides, task direction was learned dynamically using the VGG16 network without defining which is the task to be performed. Therefore, the RL-controller learns on-the-fly the guiding forces with a limited amount of training episodes. Finally, the learned policy presented acceptable behavior in terms of convergence and performance, being able to assist successfully the subjects to execute the task faster and in a personal way.

Despite the good results obtained with the implemented simulator, it was not possible to completely mimic the human behavior. In this context, modelling techniques as cybernetics modelling or system identification methods could be used to improve the simulator functioning for future training.
In addition, different RL algorithms (for instance: PPO~\cite{DBLP:journals/corr/SchulmanWDRK17}, 
etc) could be tested in simulations in order to observe resulting behavior in the implemented system and perform a comparison between them in terms of performance and required number of training episodes.

In the real system, most of the basic parameters were set empirically. For instance: proportional gains in algorithms. These could be further optimized in order to improve performance in the implemented system. In the design of the system it was intended that the slave side follows all the orders received by the master side. So that, any movement in the slave side is the result of tracking the interaction between the user and the RL-controller. Instead of using a proportional controller, a proportional-integral (PI) controller or more advanced technique could be used for this task.

Despite the reward function capturing most of the intention of the different subjects, the threshold $x$ was defined according the Subject 0 behavior. However, this value could might be not optimal for the other subjects. As such, the reward function could be improved by changing the $x$ value according the subject preference.

Finally, the use of the image of the initial state allows the users to perform tasks that depend on this kind of information, the pick-and-place task for example. However, to perform \textit{any} kind of task, for instance the peg-in-hole task, it is desirable to have information about the initial and the current state. The same holds for the task encoding bottleneck, which should be adjusted to allow the network to encode all relevant task information.

\ifCLASSOPTIONcaptionsoff
  \newpage
\fi



\bibliographystyle{IEEEtran}
\bibliography{references}
%

%







\end{document}